\documentclass[runningheads]{llncs}

\usepackage{eccv}

\usepackage{eccvabbrv}

\usepackage{graphicx}
\usepackage{booktabs}
\usepackage{multirow}
\usepackage[table]{xcolor} 
\usepackage{amssymb} 
\usepackage{pifont}  

\usepackage[accsupp]{axessibility}  

\newcommand{\cmark}{\ding{51}}
\newcommand{\xmark}{\ding{55}}
\definecolor{color_intra}{HTML}{FFF4D2}  
\definecolor{color_inter}{HTML}{D9EAD3}  
\definecolor{color_count}{HTML}{FCE5CD}  
\definecolor{color_thema}{HTML}{D0E0E3}  

\usepackage{hyperref}

\usepackage{orcidlink}

\begin{document}

\title{Video-HolmesV2: Can MLLMs Reason with Spatio-Temporal Audio-Visual Evidence in Long Videos?} 

\titlerunning{Video-HolmesV2}

\author{Zhaoyang Wei\inst{1,2}\orcidlink{0000-0002-5708-7157} \and
Zipeng Wang\inst{1}\orcidlink{0009-0006-0918-5357} \and
Yushe Cao\inst{3}
\and
Chenhui Qiang\inst{2}\orcidlink{0000-0002-5358-8659}
\and
Shuaibing Cheng\inst{1}
\and
Xuesong Yang\inst{1}
\and
Sen Nie\inst{1}
\and
Bowen Jiang\inst{1}
\and
Wenchao Ding\inst{2}
\and
Yanchao Hao\inst{2}
\and
Zheng Wei\inst{2}
\and
Xuehui Yu\inst{2}
\and
Zhenjun Han\inst{1}\thanks{Corresponding author.}
}

\authorrunning{Zhaoyang Wei et al.}

\institute{University of Chinese Academy of Sciences, China \and
Tencent, China \and
Tsinghua University, China}

\maketitle

\begin{abstract}
Multimodal Large Language Models have demonstrated impressive video understanding, yet their ability to reason over long-form narratives is often masked by visual-centric evaluations and inefficient context processing. Existing benchmarks over-rely on visual heuristics while marginalizing auditory cues, effectively reducing models to "silent observers" that bypass genuine cross-modal reasoning. Moreover, standard dense sampling creates an \textbf{evidence–context trade-off}: increasing frames to capture evidence inevitably leads to attention distraction and token explosion.
To bridge these gaps, we present \textbf{Video-HolmesV2}, a novel benchmark designed for Deep Audio-Visual Coupling. Unlike previous works, it enforces an Evidence-Based Evaluation, requiring models to justify answers with precise \textbf{spatio-temporal audio-visual evidence}, thereby reducing confounding effects of guessing and hallucinated evidence. To support this, we introduce: (1) a Multi-Model Cross-Verification pipeline to ensure task rigor; (2) a Spatio-temporal Evidence-Aware Metric for fine-grained calibration. Furthermore, we propose an Audio-Text Guided Token Compression framework. By fusing task intent with auditory anchors, our method distills high-value reasoning cues to mitigate long-context noise. In our evaluation, even strong proprietary models achieve below 60\% accuracy, while our approach outperforms over comparable open-source omni-models.
\end{abstract}

\section{Introduction}
\label{sec:intro}

The landscape of video understanding has been fundamentally transformed by the rapid evolution of Multimodal Large Language Models (MLLMs)~\cite{li2023videochat,maaz2023video,zhang2024video,chen2024internvl,wang2024qwen2}. While these models exhibit remarkable proficiency in short-clip perception, comprehending long-form narratives remains a formidable challenge~\cite{song2024moviechat,mangalam2023egoschema,wang2024lvbench}.  To address this, recent ``omni-modal'' architectures—such as GPT-4o~\cite{4o}, Gemini 2.5 Pro~\cite{gemini25pro}, and Qwen2.5-Omni~\cite{xu2025qwen2}—natively integrate auditory streams to facilitate holistic understanding of extended temporal contexts. However, evaluation paradigms have lagged significantly behind these architectural advances. Existing benchmarks—from short-clip assessments~\cite{li2024mvbench,Fang2024MMBenchVideoAL,cheng2025video,hong2025worldsense} to hour-long evaluations like Video-MME~\cite{hu2025video} and LongVideoBench~\cite{wu2024longvideobench}—remain predominantly \textit{visual-centric}, inherently relegating the auditory channel to a secondary, often neglected signal. This systematic bias reduces MLLMs to "silent observers" that overlook dynamic acoustic semantics while over-relying on visual patterns. Such visual fixation creates a critical bottleneck for long video understanding, where visual and auditory cues are deeply intertwined rather than merely additive.
\begin{table*}[t]
\centering
\resizebox{\textwidth}{!}{%
\begin{tabular}{lccccccc}
\toprule
\textbf{Benchmarks} & \textbf{Modality} & \textbf{\# Videos} & \textbf{Avg. Len} & \textbf{\# QA Pairs} & \textbf{A-V Corr.} & \textbf{S-T Evid.} & \textbf{Evid. Chain} \\ 
\midrule
\multicolumn{8}{l}{\textit{\textbf{Video-Only Benchmarks}}} \\
ActivityNet-QA~\cite{yu2019activitynet} & V & 5,800 & 180s & 58,000 & \xmark & \xmark & \xmark \\
MMBench-Video~\cite{fu2024video} & V & 609 & 165s & 1,998  & \xmark & \xmark & \xmark \\
EgoSchema~\cite{mangalam2023egoschema} & V & 5,063 & 180s & 5,063  & \xmark & \xmark & \xmark \\
Video-MME~\cite{fu2025video} & V & 900 & 1018s & 2,700 & \xmark & \xmark & \xmark \\ 
\midrule
\multicolumn{8}{l}{\textit{\textbf{Audio-Visual Benchmarks}}} \\
WorldSense~\cite{hong2025worldsense} & A+V & 1,662 & 141s & 3,172  & \cmark & \xmark & \xmark \\
OmniVideoBench~\cite{li2025omnivideobench} & A+V & 628 &  384s & 1000 & \cmark & \xmark & \xmark \\ 
\midrule
\multicolumn{8}{l}{\textit{\textbf{Our Series}}} \\
Video-Holmes (V1)~\cite{cheng2025video} & A+V & 270 & $<$300s & 1,837  & \xmark & \xmark & \cmark \\
\rowcolor{gray!15} 
\textbf{Video-HolmesV2 (Ours)} & \textbf{A+V} & \textbf{784} & \textbf{600-7500s} & \textbf{4000} & \textbf{\cmark} & \textbf{\cmark} & \textbf{\cmark} \\ 
\bottomrule
\end{tabular}%
}
\caption{Comparison with existing benchmarks. Video-HolmesV2 stands out by offering large-scale QA pairs, long-form video understanding, and unique spatio-temporal audio-visual evidence, surpassing both video-only and recent audio-visual benchmarks.}
\label{tab:comparison}
\end{table*}

\begin{figure*}[t]
  \centering
  \includegraphics[width=0.97\linewidth, height=9.6cm]{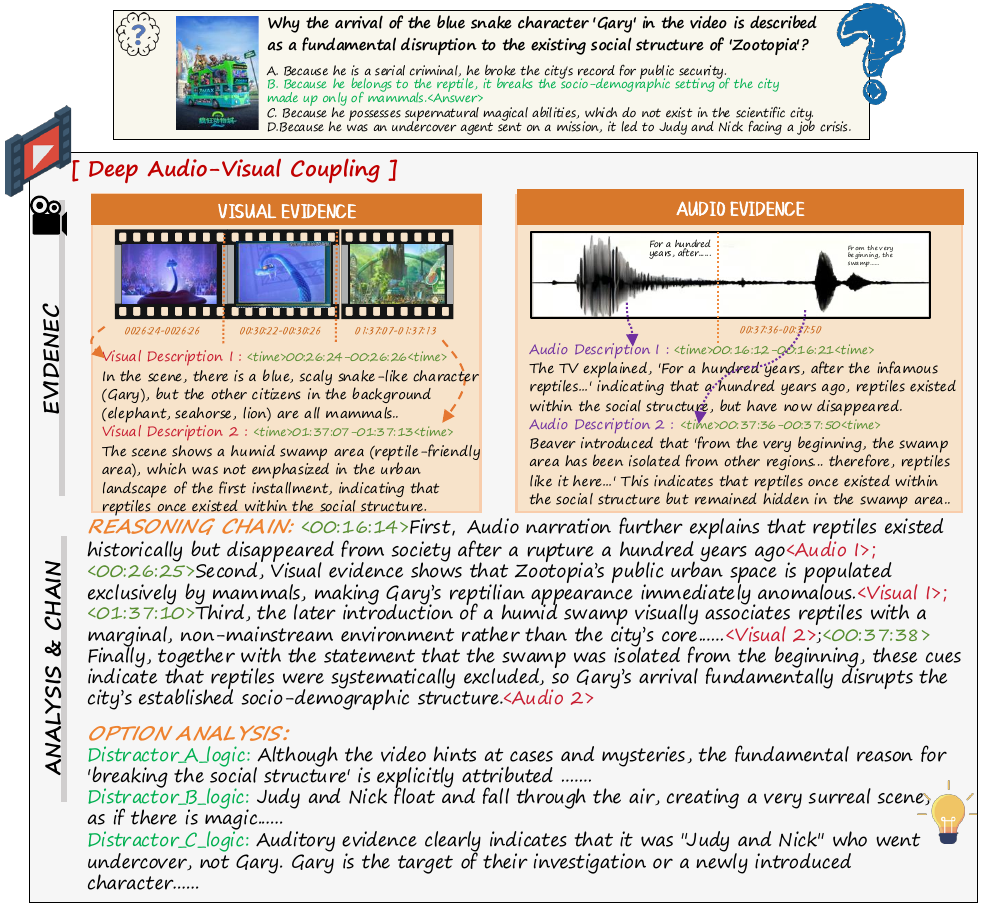} 
  
  \caption{Deep Audio-Visual Coupling in Video-HolmesV2. 
  Using \textit{Zootopia}, we demonstrate the necessity of cross-modal synthesis. Visual cues alone (identifying a blue snake) fail to imply "social disruption" without the auditory historical context (reptiles disappeared centuries ago). Uniquely, our benchmark provides an Evidence-Based Evaluation, requiring models to ground answers in precise spatio-temporal timestamps from both visual and auditory channels.}
  \label{fig:teaser}
\end{figure*}


Beyond the necessity of cross-modal coupling, traditional metrics that rely solely on answer accuracy fail to capture long-range cross-modal causality, as they cannot distinguish genuine reasoning from visually-biased hallucinations~\cite{cobbe2021training}. To bridge this critical gap, we present Video-HolmesV2, a large-scale benchmark comprising 784 long-form videos (averaging 43 minutes) specifically designed to enforce deep audio-visual coupling (A-V Corr.). As detailed in Tab.~\ref{tab:comparison}, unlike existing datasets~\cite{hu2025video,zhao2025mmvu} that prioritize visual dominance or lack intermediate supervision, Video-HolmesV2 pioneers an Evidence-Based Evaluation. This rigorous paradigm mandates that models go beyond merely predicting the correct outcome; they must actively construct a logical deduction chain (Evid. Chain) substantiated by precise, spatio-temporal audio-visual evidence (S-T Evid.), thereby proving that their answers are grounded in concrete multimodal facts.

\begin{figure*}[t!]
  \centering
  \begin{subfigure}[c]{0.48\textwidth}
    \centering
    \includegraphics[width=\linewidth]{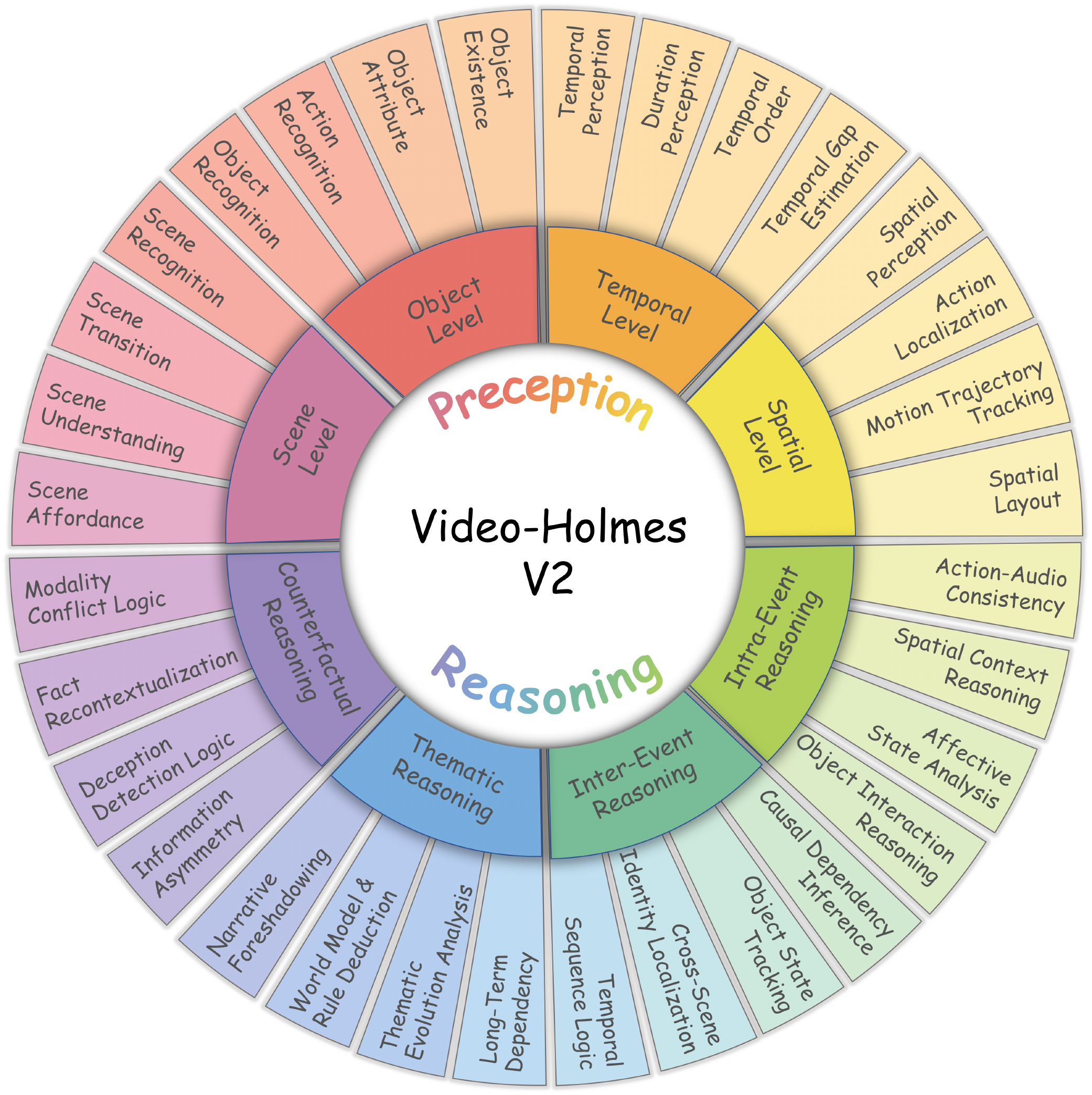} 
    \caption{Hierarchical Task Taxonomy}
    \label{fig:taxonomy}
  \end{subfigure}
  \hfill 
  \begin{subfigure}[c]{0.5\textwidth}
    \centering
    \includegraphics[width=\linewidth]{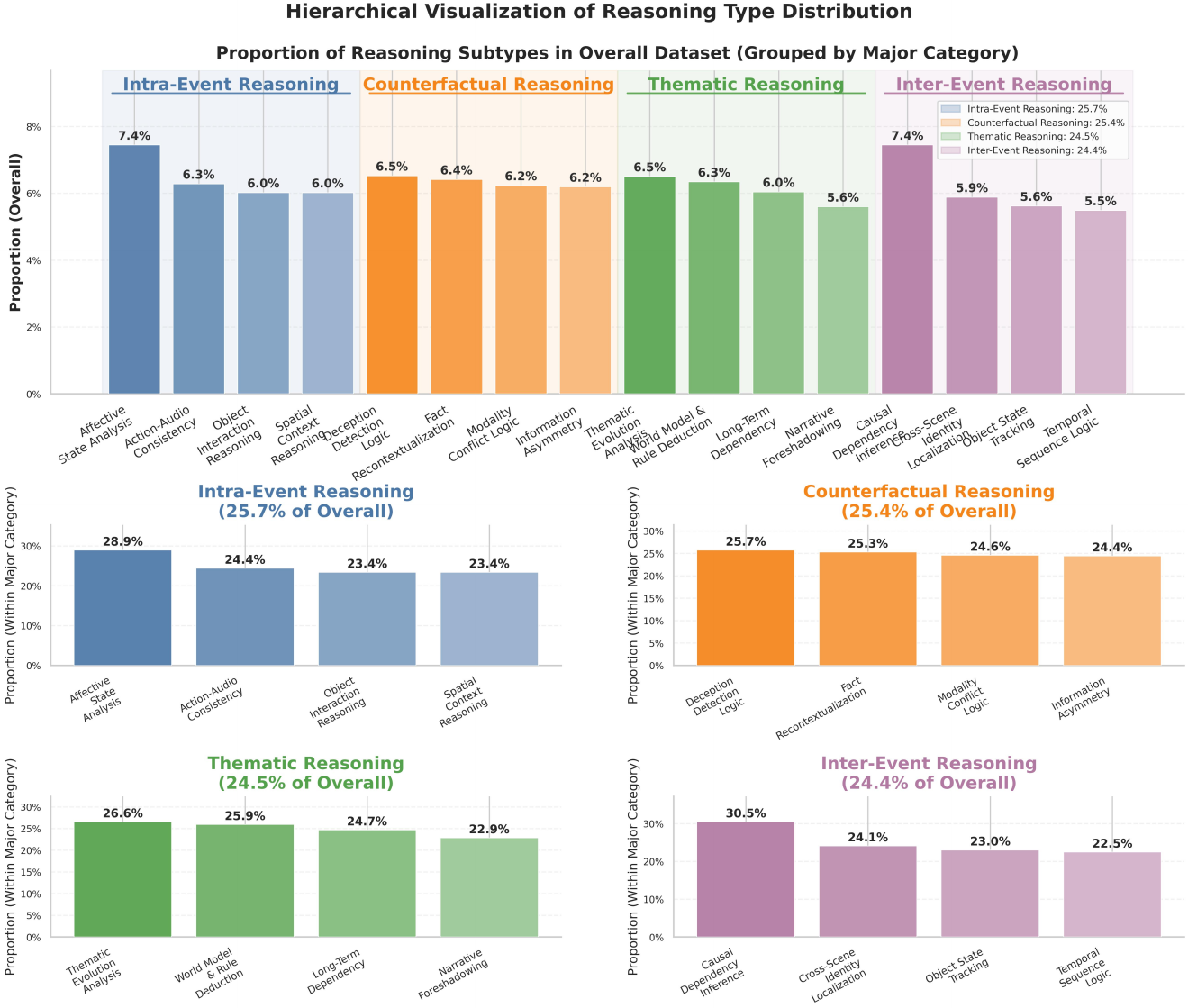} 
    \caption{Reasoning Type Distribution}
    \label{fig:distribution}
  \end{subfigure}
  
  
  \caption{The Taxonomy and Data Distribution of Video-HolmesV2. 
  (a) Our \textbf{Hierarchical Taxonomy} bridges foundational Perception (inner rings) with high-order Reasoning (outer rings), uniquely emphasizing \textit{Causal}, \textit{Counterfactual}, and \textit{Thematic} logic. 
  (b) The \textbf{Statistical Distribution} validates this design, revealing a balanced composition across four major reasoning pillars ($\approx$ 25\% each), ensuring the benchmark evaluates diverse cognitive capabilities without category bias.}
  \label{fig:taxonomy_and_stats}
\end{figure*}

To ensure uncompromised data quality, Video-HolmesV2 is curated from recent high-fidelity cinematography and meticulously annotated via a multi-model cross-verification pipeline to eliminate unimodal biases. Consequently, our task taxonomy transcends simple perception to challenge high-order cognition, including causal and counterfactual deduction (Fig.~\ref{fig:taxonomy}). A representative example of this intricate coupling is illustrated in Fig.~\ref{fig:teaser}: visually identifying a ``blue snake'' only implies a social threat when synthesized with the auditory context that reptiles vanished centuries ago. Furthermore, to robustly assess these complex reasoning chains, we introduce a Spatio-Temporal Evidence-Aware Evaluation. This metric replaces rigid thresholds with Gaussian-based soft scoring and strictly conditions the final assessment on valid evidence extraction, effectively disentangling genuine multimodal reasoning from spurious guessing.

To effectively navigate the intricate causal evidence in ultra-long videos, high-density frame sampling is a prerequisite to capture fleeting evidence. However, this precipitates a severe token explosion that overwhelms the context windows of current MLLMs. While token compression offers a remedy, prevailing strategies often rely on naive top-K truncation or unimodal guidance, inadvertently severing vital cross-modal links or discarding macroscopic context. To resolve this efficiency-accuracy dilemma, we propose a \textbf{Audio-Text Guided Token Compression} framework. By fusing text-guided subjective intent with audio-guided objective events, our approach comprehensively evaluates visual token relevance. Instead of rigid pruning, it performs fine-grained deduplication on redundant high-scoring patches and dynamically reallocates the saved budget to retain global context, ensuring both microscopic evidence and narrative atmosphere are preserved. Extensive experiments demonstrate that this paradigm not only excels on Video-HolmesV2 but also achieves state-of-the-art performance across multiple established video understanding benchmarks.

Our main contributions are summarized as follows:

\begin{itemize}
\item We introduce Video-HolmesV2, a benchmark of 784 long-form videos featuring \textit{Deep Audio-Visual Coupling}. It challenges models to resolve complex causal dependencies relying on inextricably linked visual and auditory clues.
    
\item We curate precise Spatio-Temporal Audio-Visual Evidence via multi-model cross-verification, enabling both high-fidelity bimodal annotations and a novel evaluation metric that strictly conditions reasoning accuracy on valid temporal grounding, effectively penalizing ungrounded speculation.
    
\item We propose an Audio-Text Guided Token Compression framework. By combining bimodal scoring with differentiated deduplication, it preserves critical evidence and achieves superior reasoning accuracy compared to existing token compression methods on long-video tasks.
\end{itemize}

\section{Related Works}

\subsection{Multimodal Models for Video Understanding}
The field of video understanding has rapidly evolved with Multimodal Large Language Models (MLLMs). Initially, models treated videos as sequences of images~\cite{li2023videochat,maaz2023video,zhang2024video}, leveraging image-based foundations like CLIP~\cite{radford2021CLIP} and modular encoders to perform unified inference~\cite{chen2024internvl,zhu2025internvl3,wang2024qwen2}. Recently, the frontier has shifted towards ``Omni-modal'' architectures capable of natively processing interwoven data streams. Proprietary models like GPT-4o~\cite{4o} and Gemini 2.5 Pro~\cite{gemini25pro}, alongside open-source efforts such as Qwen2.5-Omni~\cite{xu2025qwen2}, VITA~\cite{vita,vita-1.5}, Video-SALMONN~\cite{video-salmonn}, and InteractiveOmni~\cite{tong2025interactiveomni}, have explored end-to-end capabilities for simultaneous audio-video understanding. Despite these architectural advancements, evaluating and processing thousands of frames introduces severe computational burdens. While recent methods have explored token pruning~\cite{shao2025tokenstalkmuchsurvey,flexivideo}, keyframe selection~\cite{wang2025videotreeadaptivetreebasedvideo}, and memory-augmented frameworks~\cite{liu2025videomindchainofloraagentlong} to alleviate this token explosion, efficiently bridging the gap between multimodal inputs and complex reasoning in long contexts remains a significant challenge.

\subsection{General and Long Video Understanding Benchmarks}
Benchmarks act as the compass for model development. Early evaluations primarily assessed fundamental perception capabilities on short clips~\cite{xu2017video,zhong2022video,yu2019activitynet,longVE,grip}. As models progressed, the community introduced benchmarks targeting sophisticated reasoning, comprehensive short-clip abilities, Chain-of-Thought (CoT), and academic domain knowledge~\cite{li2024mvbench,Fang2024MMBenchVideoAL,cheng2025video,hong2025worldsense,zhao2025mmvu,qi2025vcr,SAPNet}. To assess temporal coherence and memory retention, datasets like MovieChat-1K~\cite{song2024moviechat}, EgoSchema~\cite{mangalam2023egoschema}, LVBench~\cite{wang2024lvbench}, Video-MME~\cite{hu2025video}, and LongVideoBench~\cite{wu2024longvideobench} have extended evaluations to hour-long sequences. Concurrently, in the realm of multi-modal perception, benchmarks like AVQA~\cite{yang2022avqa} and Music-AVQA~\cite{Li2022Learning} have integrated audio cues. However, existing long-video benchmarks often treat audio as a secondary modality, while audio-visual datasets are largely confined to short durations. More importantly, they do not strictly require models to substantiate their answers with precise spatio-temporal evidence, a gap highlighted by recent calls for process-level annotation~\cite{cobbe2021training, longVE, verbench, adbench}. Video-HolmesV2 directly addresses this by introducing a benchmark centered on omni-modal, event-based reasoning over truly long durations.

\section{Benchmark Construction.}
\subsection{Initial Video Collection and Technical Curation}

Our data construction begins by collecting 3704 high-quality, post-2024 cinematic videos (ranging from 10 to 150 minutes) to evaluate genuine zero-shot capabilities and prevent data contamination. We prioritize visually and acoustically rich narratives (e.g., award-winning shorts, episodic series, movies) while systematically excluding unimodal-dominant or causally sparse genres, such as news reports and sports matches. Detailed duration stratification and engagement thresholds are provided in Appendix~\ref{sec:app_video_collection}.
\paragraph{\textbf{Multi-Dimensional Analyzability Assessment.}} 
To rigorously quantify the multimodal reasoning potential of the filtered candidates, we employ a consortium of state-of-the-art native MLLMs (e.g., Gemini 3 Pro, GPT-5.2, Qwen3-VL-235B-A22B-thinking) to evaluate each video. Specifically, videos are scored on a scale from 1 to 10 across five core dimensions:\\
\textbf{1. Audio-Visual Complementarity:} The strict necessity of synthesizing both visual and auditory channels, ensuring neither modality is sufficient in isolation.\\
\textbf{2. Cross-Modal Causal Complexity:} The depth of multi-hop cause-and-effect chains that inextricably span across modalities and time.\\
\textbf{3. Dynamic Character Arc:} The meaningful evolution of key entities over time, evidenced concurrently by visual changes and auditory shifts.\\
\textbf{4. Narrative Coherence:} The structural integrity and logical consistency of the storyline, filtering out videos with disjointed editing or fallacies.\\
\textbf{5. Audience Critical Reception:} A meta-metric gauging narrative depth based on whether the video provokes high-quality intellectual discussion.
A comprehensive elaboration of these criteria and the hierarchical processing strategy for ultra-long sequences can be found in Appendix~\ref{sec:app_video_collection}.
\paragraph{\textbf{Quantitative Curation and Stratification.}}
To synthesize the multi-dimensional assessments into a robust decision, we implement a rigorous three-stage quantitative pipeline. This process filters low-quality content, enforces strict multimodal dependencies, and identifies high-difficulty samples based on model uncertainty.

\noindent\textbf{1. Bias Mitigation via Z-Score Normalization.} 
Different VLMs exhibit intrinsic scoring biases (e.g., varying scales of leniency). To mitigate this, we standardize the raw score $x_{d,i}$ of model $i$ on dimension $d$ to a Z-score via $z_{d,i} = (x_{d,i} - \mu_{i}) / \sigma_{i}$. This step aligns the evaluator distributions, ensuring the selection is driven by relative ranking rather than absolute score artifacts.

\noindent\textbf{2. Priority-Weighted Aggregation with Modality Veto.} 
We derive a unified Analyzability Score ($S_{\text{final}}$) via a weighted aggregation of the normalized dimensions. We assign higher importance weights ($w_d$) to Audio-Visual Complementarity and Cross-Modal Causal Complexity to emphasize reasoning:
{\scriptsize
\begin{equation}
S_{\text{final}} = \sum_{d \in \mathcal{D}} w_d \cdot \left( \frac{1}{N} \sum_{i=1}^{N} z_{d,i} \right).
\end{equation}}
Crucially, to prevent the inclusion of ``high-quality but single-modality'' videos, we enforce a \textbf{Modality Veto}. Regardless of the total $S_{\text{final}}$, any video scoring below a specific threshold ($z < -0.5$) in the \textit{Audio-Visual Complementarity} dimension is automatically discarded. 

\noindent\textbf{3. Uncertainty-Driven Data Stratification.} 
To quantify evaluator consensus and highlight narrative ambiguity, we define a \textbf{Divergence Score ($D_{\text{video}}$)} as the arithmetic mean of the standard deviations across all dimensions:
{\scriptsize
\begin{equation}
D_{\text{video}} = \frac{1}{|\mathcal{D}|} \sum_{d \in \mathcal{D}} \sqrt{\frac{\sum_{i=1}^{N} (z_{d,i} - \bar{z}_d)^2}{N}}.
\end{equation}}
A high $D_{\text{video}}$ indicates significant SOTA model disagreement. Instead of a fixed arbitrary threshold, we define an adaptive high-divergence cutoff $\tau_{div} = \mu_{D} + \sigma_{D}$ to statistically flag the top $\approx 16\%$ most controversial samples. Consequently, videos are strictly categorized into: (1) \textbf{Core Set} ($S_{\text{final}} \ge 0, D_{\text{video}} < \tau_{div}$); (2) \textbf{Rejected} ($S_{\text{final}} < 0$ or Vetoed); and (3) \textbf{Hard Candidates} ($D_{\text{video}} \ge \tau_{div}$). These high-variance candidates are directly routed to expert human review for further adjudication.

\begin{figure*}[t!]
  \centering
  \includegraphics[width=1.0\linewidth]{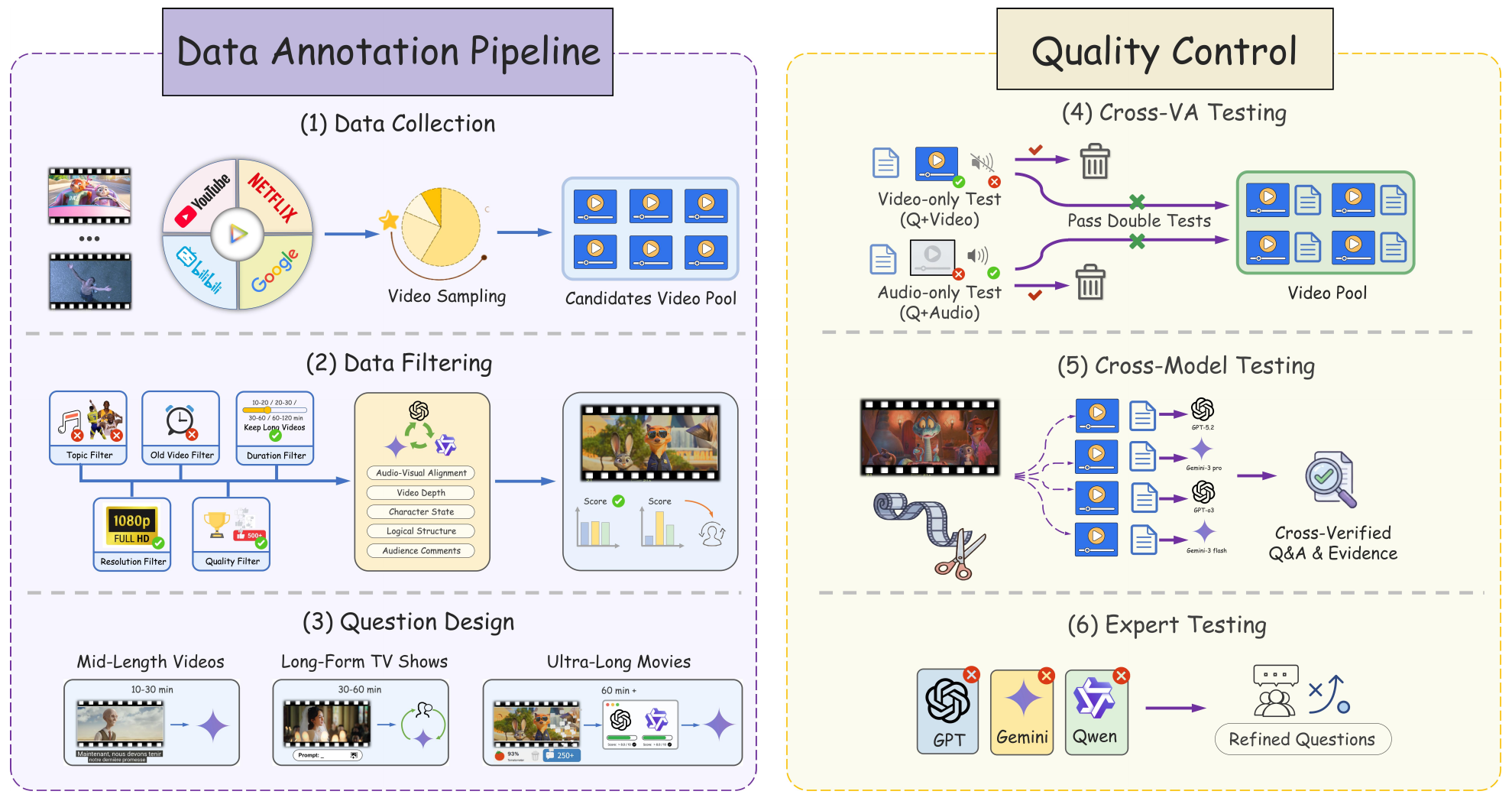} 
  \caption{\textbf{The Scalable Annotation and Quality Control Pipeline.} 
  Our construction process integrates native long-context MLLM generation with rigorous human-in-the-loop verification. The pipeline moves from (1-3) diverse sourcing and stratified question generation to a strict (4-6) multi-stage validation protocol. A critical innovation is step (4) \textit{Cross-VA Testing}, which acts as a ``Unimodal Trap'' to automatically discard questions solvable by audio or vision alone, ensuring that every sample in Video-HolmesV2 strictly enforces multimodal interdependence. Finally, expert review (6) adjudicates hard negatives to guarantee logical determinism.}
  \label{fig:pipeline}
\end{figure*}
\subsection{Scalable Annotation via Native Long-Context Understanding}
\label{sec:annotation_pipeline}

As illustrated in Fig.~\ref{fig:pipeline}, we propose a unified annotation pipeline leveraging native long-context MLLMs. Unlike traditional clip-based methods, we process complete narrative streams to capture long-range dependencies. The workflow comprises four rigorous phases, with detailed process provided in Appendix~\ref{sec:app_annotation}.

\noindent\textbf{Phase 1: Stratified Question Generation} (Fig.~\ref{fig:pipeline}, Step 3)\textbf{.} 
Recognizing varying narrative complexities, we employ a length-differentiated strategy: direct \textit{Constraint-Based Prompting} for short films, \textit{Contextual Priming} via injected lore for episodic series, and \textit{Comment-Driven Inverse Reasoning} for feature films. Crucially, the models are constrained to output not only the complex QA pairs but also explicit grounded \textit{spatio-temporal audio-visual evidence}.

\noindent\textbf{Phase 2: The ``Unimodal Trap'' Blind Filtration} (Fig.~\ref{fig:pipeline}, Step 4)\textbf{.} 
To strictly enforce multimodal reasoning, we introduce an adversarial filter. Generated QA pairs are evaluated against ablated \textit{Visual-Only} (silent frames) and \textit{Audio-Only} (subtitles) inputs. Any question solvable by a single modality with high confidence is automatically discarded as a ``pseudo-multimodal'' instance.

\noindent\textbf{Phase 3: Segment-Level Cross-Verification} (Fig.~\ref{fig:pipeline}, Step 5)\textbf{.} 
We isolate the specific narrative segments containing the proposed evidence and deploy a diverse consortium of reasoning-heavy SOTA models as validators. A sample is retained only if a majority consensus is reached on both the correct answer and the precise evidence timestamps (Intersection over Union $> 0.75$).

\noindent\textbf{Phase 4: Expert Adjudication for Hard Negatives} (Fig.~\ref{fig:pipeline}, Step 6)\textbf{.} 
Samples where the automated ensemble disagrees or fails are routed to a Double-Blind Expert Review. Specialists independently assess whether the failure stems from model hallucination or genuine, deep logical complexity. Validated samples from this queue are salvaged to form the \textit{Hard Subset} of our benchmark, representing the upper bound of current AI reasoning capabilities.

\section{Evaluation Metrics}
We introduce a comprehensive evaluation protocol that moves beyond rigid thresholding to jointly assess multiple-choice reasoning accuracy and spatio-temporal grounding quality in long videos.

\noindent{\textbf{4.1 Answer Accuracy ($Acc_{\text{MCQ}}$)}.}
The correctness of the reasoning conclusion is binary. For a chosen option $\hat{y}$ and ground truth $y$:
{\footnotesize
\begin{equation}
    Acc_{\text{MCQ}} = \mathbb{I}(\hat{y} = y)
\end{equation}}
\noindent\textbf{4.2 Evidence Quality ($S_{\text{evidence}}$).}
Measuring evidence quality in long videos is challenging due to variable event durations. We propose a robust metric that combines length-adaptive temporal alignment with semantic matching.

\textit{1. Temporal Soft Alignment ($S_{\text{temp}}$).}
To accommodate systematic temporal drift ($\tau_{drift}={avg \ duration}*0.05$) while severely penalizing duration mismatch, we formulate a soft score that multiplies a length-similarity ratio by a Gaussian decay. For a predicted interval $p$ and ground-truth $g$ with lengths $L_{(\cdot)}$ and centers $C(\cdot)$:
{\scriptsize
\begin{equation}
    S_{\text{raw}}(p, g) = \left( \frac{\min(L_p, L_g)}{\max(L_p, L_g)} \right) \cdot \exp\left( - \frac{\max(0, \| C(p) - C(g) \| - \tau_{drift})^2}{2 \cdot (\alpha_{gauss} \cdot L_g)^2} \right)
\end{equation}}
where $\alpha_{gauss}=0.5$. The Gaussian factor elegantly absorbs minor localization shifts within $\tau_{drift}$, while the leading ratio strictly enforces structural consistency. To handle fragmented outputs, we coalesce proximal predictions (gap $\le \tau$) into unified spans $\mathcal{P}_c$. Following bipartite matching with ground-truths, the temporal score is strictly prediction-centric: $S_{\text{temp}} = \frac{1}{|\mathcal{P}_c|} \sum_{p \in \mathcal{P}_c} S_{\text{raw}}(p, g^*)$. Unmatched predictions score zero to penalize hallucinations. Crucially, since logical deduction often requires only partial clues, unretrieved ground-truths are not penalized here; this recall penalty is inherently deferred to $S_{\text{sem}}$.

\textit{2. Semantic Matching ($S_{\text{sem}}$).}
Beyond mere temporal overlap, models must accurately identify causal events. Thus, we utilize Qwen3-235B-A22B-Thinking to evaluate semantic equivalence exclusively for temporally intersecting predictions ($S_{\text{temp}}>0$). 
Since complex reasoning often requires only a subset of clues, we apply the $F_{0.5}$ score (weighting precision twice over recall) to heavily penalize hallucinations while tolerating partial, sufficient evidence:
{\scriptsize
\begin{equation}
    S_{\text{sem}} = (1 + 0.5^2) \cdot \frac{P_{\text{sem}} \cdot R_{\text{sem}}}{0.5^2 \cdot P_{\text{sem}} + R_{\text{sem}}}
\end{equation}}
\noindent{\textbf{4.3 Overall Performance ($S_{\text{total}}$).}}
The final score enforces a strict dependency between the answer and evidence. We first combine the temporal and semantic scores into a unified evidence metric:
{\footnotesize
\begin{equation}
    S_{\text{evidence}} = \alpha_{evi} \cdot S_{\text{temp}} + (1 - \alpha_{evi}) \cdot S_{\text{sem}}
\end{equation}}
where $\alpha_{evi}=0.5$ balances temporal precision and semantic correctness. The total performance score is then defined as:
{\footnotesize
\begin{equation}
    S_{\text{total}} = \lambda_1 \cdot Acc_{\text{MCQ}} + \lambda_2 \cdot (Acc_{\text{MCQ}} \cdot S_{\text{evidence}})
\end{equation}}
We strictly set weights $\lambda_1 = 0.6$ and $\lambda_2 = 0.4$. The conditional term $(Acc_{\text{MCQ}} \cdot S_{\text{evidence}})$ ensures evidence quality is only rewarded when the final answer is correct, effectively penalizing ungrounded guessing.
\section{Method}
\noindent{\textbf{5.1 Lightweight Text-Guided Visual Scoring.}}
\label{sec:text_guided_scoring}
Long-form video understanding entails processing an exorbitant number of visual tokens, rendering the direct application of massive MLLM computationally prohibitive. To efficiently identify query-relevant visual signals before heavy autoregressive decoding, we introduce a lightweight, rank-supervised token selector (Qwen2.5 0.5B) distilled from a large-scale teacher model (Qwen2.5-VL 72B).

Given a textual query $q$ and a sequence of visual tokens $\mathbf{V} = \{v_1, v_2, \dots, v_N\}$, our goal is to compute a text-guided relevance score $S_{text} \in \mathbb{R}^N$. Prior studies indicate that intermediate transformer layers in VideoLLMs exhibit the highest semantic alignment between text queries and visual features. Therefore, we extract the cross-modal attention from a designated reference layer $L_{ref}$ of the teacher model. For the $i$-th visual token, the teacher's semantic relevance score $r_i^{ref}$ is defined by averaging the attention weights across all $H$ heads:
{
\scriptsize
\begin{equation}
    r_i^{ref} = \frac{1}{H} \sum_{h=1}^{H} \mathbf{A}_{q \to v_i}^{(L_{ref}, h)}
\end{equation}}
where $\mathbf{A}_{q \to v_i}^{(L_{ref}, h)}$ denotes the attention weight from the query tokens to $v_i$.
\begin{figure*}[t]
  \centering
  \includegraphics[width=1.0\textwidth]{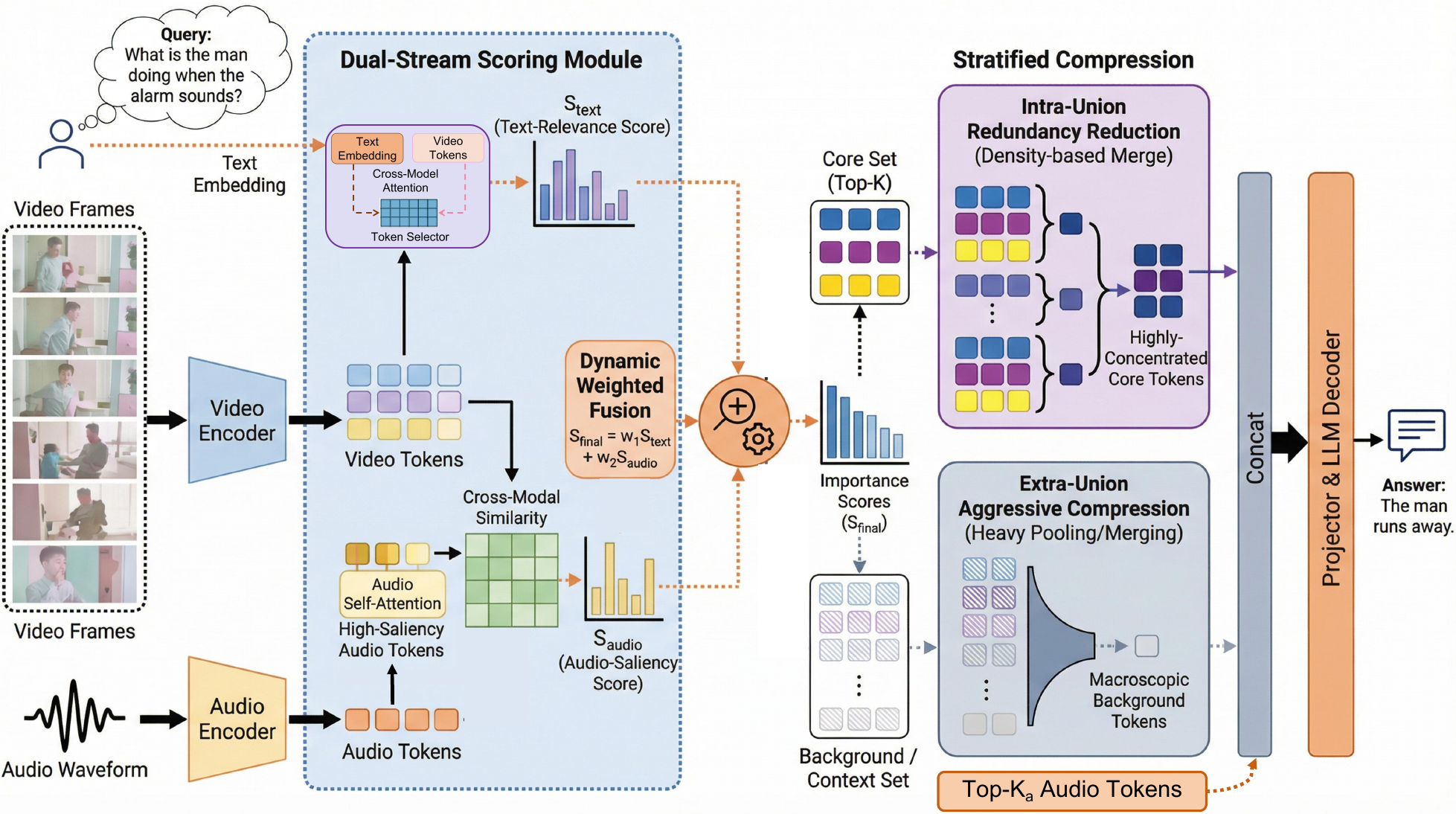}
  \caption{{The Audio-Text Guided Token Compression Framework.} 
  We fuse text-guided subjective intent ($S_{\text{text}}$) with audio-guided objective cues ($S_{\text{audio}}$) to dynamically assess visual importance. 
  A {Stratified Compression} strategy then preserves fine-grained evidence in the Core Set while aggressively pooling the Background Set to maximize context efficiency.}
  \label{fig:method_pipeline}
\end{figure*}
To circumvent the prohibitive computational cost of the teacher model during inference, we train a shallow, lightweight selector to predict these relevance scores, denoted as $\hat{\mathbf{r}} = [\hat{r}_1, \dots, \hat{r}_N]$. Instead of minimizing the absolute mean squared error, we optimize the ranking consistency between the student and the teacher using a Spearman Rank Correlation Loss. The training objective is:
{
\scriptsize
\begin{equation}
    \mathcal{L}_{rank} = 1 - \frac{\sum_{i=1}^{N} (R(r_i^{ref}) - \bar{R}^{ref}) (R(\hat{r}_i) - \bar{\hat{R}})}{\sqrt{\sum_{i=1}^{N} (R(r_i^{ref}) - \bar{R}^{ref})^2 \sum_{i=1}^{N} (R(\hat{r}_i) - \bar{\hat{R}})^2}}
\end{equation}
}
where $R(\cdot)$ denotes the continuous approximation of the ranking function via differentiable sorting, and $\bar{R}$ is the mean rank. Once trained, this lightweight selector efficiently outputs the text-guided visual score $S_{text} = \hat{\mathbf{r}}$, serving as the top-down semantic prior for our subsequent dual-stream compression. More details in Appendix~\ref{text_scoring_details}.

\noindent{\textbf{5.2 Training-Free Audio-Guided Visual Scoring.}}
\label{sec:audio_guided_scoring}
To capture crucial out-of-view or transient auditory events often omitted by text queries, we introduce a complementary, training-free audio-guided scoring mechanism that anchors visual tokens to salient acoustic cues.

The process consists of two stages: intra-modal audio filtering and cross-modal saliency propagation. Given the raw audio tokens $\mathbf{A} \in \mathbb{R}^{N_a \times d}$ extracted by the audio encoder, we first identify the most informative audio anchors. Since continuous background noise provides marginal semantic value, we leverage the intrinsic self-attention matrix $\mathbf{M} \in \mathbb{R}^{N_a \times N_a}$ from the final layer of the audio encoder. The saliency of the $j$-th audio token is quantified by the average attention it receives from all other audio tokens:
{\scriptsize
\begin{equation}
    s_a^{(j)} = \frac{1}{N_a} \sum_{k=1}^{N_a} \mathbf{M}_{k, j}
\end{equation}}
We select the top-$K_a$ (e.g., $K_a = 2048$) tokens with the highest $s_a$ to form a compact set of salient audio anchors $\hat{\mathbf{A}} \in \mathbb{R}^{K_a \times d}$.

Subsequently, we project this acoustic saliency into the visual domain. For each visual token $v_i \in \mathbf{V}$, we compute its cross-modal semantic alignment with the salient audio anchors using normalized cosine similarity. The audio-guided visual score $S_{audio}^{(i)}$ is determined by its maximum similarity to any of the audio anchors:
{\scriptsize
\begin{equation}
    S_{audio}^{(i)} = \max_{1 \le j \le K_a} \left( \frac{v_i \cdot \hat{a}_j}{\|v_i\|_2 \|\hat{a}_j\|_2} \right)
\end{equation}}
where $\hat{a}_j \in \hat{\mathbf{A}}$. By employing the max-pooling operation over the audio dimension, $S_{audio} \in \mathbb{R}^{N_v}$ effectively highlights specific visual regions that synchronously align with transient auditory peaks, serving as an indispensable counterpart to the text-guided scores. More details in Appendix~\ref{audio_scoring_details}.

\noindent{\textbf{5.3 Dual-Stream Fusion and Differentiated Compression.}}
\label{sec:fusion_and_compression}
The text-guided score $S_{text}$ and audio-guided score $S_{audio}$ capture subjective intents and objective transient events, respectively. To construct a holistic importance metric, we perform normalization on both scores, denoted as $\tilde{S}_{text}$ and $\tilde{S}_{audio}$, and compute the final fused score:
{\scriptsize
\begin{equation}
    S_{final} = \alpha_{fuse} \cdot \tilde{S}_{text} + (1 - \alpha_{fuse}) \cdot \tilde{S}_{audio}
\end{equation}}
where $\alpha_{fuse} \in [0, 1]$ balances the bimodal influences (empirically set to 0.7). 

Given a strict sequence length budget $B_{vis}$ (e.g., $B_{vis}=8192$) for the LLM, a naive Top-$B_{vis}$ truncation based on $S_{final}$ suffers from high semantic redundancy and a complete loss of macro-context. To address this, we propose a Masked Differentiated Compression strategy. We first generate a binary spatial-temporal mask $\mathcal{M} \in \{0, 1\}^{T \times H \times W}$ by thresholding the Top-$B_{vis}$ tokens of $S_{final}$. Tokens with $\mathcal{M}=1$ constitute the \textbf{Core Set} $\mathcal{C}$, while those with $\mathcal{M}=0$ form the \textbf{Background Set} $\mathcal{B}$.

\textbf{Intra-Set Fine-Grained Compression.} 
The Core Set $\mathcal{C}$ contains semantically dense but visually redundant tokens (e.g., homogeneous sky patches scoring high). We apply a masked spatio-temporal merging operation exclusively within $\mathcal{C}$. For spatially or temporally adjacent tokens $v_i, v_j \in \mathcal{C}$, if their visual feature cosine similarity exceeds a predefined threshold $\tau_{intra}$, they are merged into a single token. The fused visual feature is computed via average pooling, while its importance score inherits the maximum value: $S_{final}^{(new)} = \max(S_{final}^{(i)}, S_{final}^{(j)})$. This ensures that visual fidelity is preserved without diluting the semantic saliency. This fine-grained deduplication reduces the size of $\mathcal{C}$ from $B_{vis}$ to $n$ ($n \le B_{vis}$), dynamically freeing up a token quota $R = B_{vis} - n$.

\textbf{Extra-Set Aggressive Scavenging.} 
The liberated quota $R$ is utilized to rescue discarded tokens from the Background Set $\mathcal{B}$, reinstating macro-context. If $R > 0$, we retrieve the top $M \times R$ tokens from $\mathcal{B}$ based on $S_{final}$, where $M$ is the aggressive compression ratio (e.g., $M=4$). Since background elements primarily provide global atmosphere rather than fine-grained details, we perform aggressive $M\text{:}1$ pooling on these tokens to strictly fit them into the $R$ budget.

\textbf{Sequence Reassembly.} 
Finally, the $n$ highly-concentrated core tokens and the $R$ macroscopic background tokens are concatenated. By strictly preserving their original spatio-temporal positional embeddings during merging, the reconstructed $B_{vis}$-token and top-$K_a$ audio token sequence retains integral causal and spatial structures for the LLM's autoregressive decoding. More details in Appendix~\ref{compression_details}.

\section{Experiment}
\subsection{Experimental Settings}

\noindent\textbf{Evaluation Setup.} 
To establish a comprehensive baseline, we evaluate both closed-source frontier models (GPT-4o, Gemini) and recent open-source models (Qwen series). The open-source evaluations are deployed using the MS-SWIFT framework on 2 $\times$ H20 GPUs. For generation hyperparameters, we set the {temperature} to $0.01$ to ensure deterministic reasoning, and crucially expand max tokens to $8192$. This extended context allows models sufficient generation capacity to output detailed visual and acoustic evidence descriptions without truncation.

\noindent\textbf{Training Implementation.} 
For our token compression method, we evaluate on standard long-video benchmarks (e.g., VideoMME, WorldScene), strictly aligning our evaluation protocol with OmniZip~\cite{tao2025omnizip} for fair comparison. Our token selector is trained on 4 $\times$ H20 GPUs with a total batch size of $8$ and a learning rate of $1 \times 10^{-5}$. The training data comprises $50$K long-duration samples, randomly selected from the longest videos within the LLaVA-Video-178k dataset.
\definecolor{color_perc}{HTML}{E8F0FE}    
\definecolor{color_reason}{HTML}{FFF4D2}  

\begin{table*}[t]
\centering
\setlength{\tabcolsep}{4pt} 
\renewcommand{\arraystretch}{1.2} 
\resizebox{1\textwidth}{!}{%
\begin{tabular}{l | c c c c c | c c c c c | c}
\toprule

\multirow{2}{*}{\textbf{Model}} & 
\multicolumn{5}{c|}{\cellcolor{color_perc}\textbf{Perception}} & 
\multicolumn{5}{c|}{\cellcolor{color_reason}\textbf{Reasoning}} & 
\multirow{2}{*}{\textbf{Overall}} \\

& 
\textit{Obj.} & \textit{Scn.} & \textit{Spa.} & \textit{Tem.} & \textbf{Ave.} & 
\textit{Intra} & \textit{Inter} & \textit{Count.} & \textit{Them.} & \textbf{Ave.} & \\
\midrule

\multicolumn{12}{l}{\textit{\textbf{Proprietary Models}}} \\
\hline
GPT-4o
& 52.1 & 60.2 & 65.4 & 46.7 & 56.1 
& 43.5 & 40.2 & 50.1 & 51.8 & 46.4 
& 46.9 \\                            

Gemini-2.5-Flash
& 63.5 & 68.2 & 71.4 & 64.9 & 67.0 
& 63.5 & 62.8 & 66.4 & 68.1 & 65.2 
& 65.3 \\                            

Gemini-2.5-Pro
& \textbf{74.3} & \textbf{77.3} & \textbf{81.8} & \textbf{70.6} & \textbf{75.8} 
& \textbf{71.5} & \textbf{70.8} & \textbf{74.6} & \textbf{75.5} & \textbf{73.1} 
& \textbf{73.2} \\                            

\midrule
\multicolumn{12}{l}{\textit{\textbf{Open Source Models}}} \\
\hline

Qwen3-VL-4B 
& 38.7 & 43.6 & 72.3 & 33.3 & 47.3 
& 25.0 & 20.7 & 28.3 & 34.2 & 27.0 
& 28.0 \\

Qwen2.5-VL-7B 
& 30.7 & 56.4 & 46.8 & 38.9 & 41.9
& 37.3 & 31.2 & 40.7 & 42.7 & 37.9
& 38.1 \\

Qwen3-VL-8B 
& 33.8 & 56.4 & 75.1 & 35.2 & 49.5
& 37.3 & 31.3 & 41.6 & 46.2 & 39.0
& 39.5 \\


Qwen2.5-VL-32B 
& 43.6 & 53.9 & 70.2 & 44.4 & 52.7
& 41.0 & 37.1 & 46.7 & 47.9 & 43.2
& 43.6 \\


MiniCPM-V-4.5 
& 38.7 & 23.1 & 51.1 & 38.9 & 38.6
& 43.8 & 43.0 & 48.7 & 54.3 & 47.4
& 47.0 \\

InternVL3-8B 
& 48.2 & 52.5 & 60.4 & 39.5 & 50.6
& 44.3 & 43.5 & 51.1 & 53.8 & 48.0
& 48.1 \\

Qwen3-VL-30B-A3B 
& 49.5 & 58.2 & 62.8 & 45.2 & 53.7
& 42.2 & 41.8 & 52.2 & 56.1 & 48.1
& 48.4 \\

Qwen3-VL-32B 
& 43.1 & 66.5 & 73.2 & 37.8 & 54.4
& 44.3 & 42.1 & 55.3 & 55.8 & 49.4
& 49.6 \\

Qwen2.5-VL-72B 
& 43.6 & 51.3 & 72.3 & 41.7 & 52.2
& 44.8 & 42.7 & 56.5 & 57.3 & 50.1
& 50.2 \\

Qwen3-VL-235B-A22B 
& \textbf{54.8} & \textbf{67.2} & \textbf{75.9} & \textbf{48.4} & \textbf{58.5}
& \textbf{48.5} & \textbf{48.5} & \textbf{60.4} & \textbf{63.7} & \textbf{55.1}
& \textbf{55.3} \\

\hline
Qwen2.5-Omni-7B 
& 33.9 & 59.0 & 53.2 & 27.8 & 42.9
& 40.7 & 35.6 & 43.0 & 43.1 & 40.6
& 40.7 \\

\textbf{Ours} 
& 37.1 & 66.7 & 59.6 & 25.0 & 46.7 
& 42.7 & 38.4 & 47.7 & 48.8 & 44.4 
& 44.5 {\color{red}$(\uparrow 3.8)$} \\     


\bottomrule
\end{tabular}
}
\caption{Comprehensive Benchmark Results for Perception and Reasoning. The final \textbf{Overall} score is strictly weighted by the sample size of each module. Models are sorted by their final weighted Overall score within their categories. The best scores among proprietary and open-source models are highlighted in \textbf{bold}, respectively.}
\label{tab:perc_reason_benchmark}
\end{table*}
\definecolor{color_reason}{HTML}{FFF4D2}  
\definecolor{color_evid}{HTML}{E4D7F5}    

\begin{table*}[t]
\centering
\setlength{\tabcolsep}{5.5pt} 
\renewcommand{\arraystretch}{1.2} 
\resizebox{1\textwidth}{!}{%
\begin{tabular}{l | c c c c c | c c c c c | c}
\toprule

\multirow{2}{*}{\textbf{Model}} & 
\multicolumn{5}{c|}{\cellcolor{color_reason}\textbf{Reasoning (MCQ Accuracy \%)}} & 
\multicolumn{5}{c|}{\cellcolor{color_evid}\textbf{Evidence (Score)}} & 
\multirow{2}{*}{\textbf{Overall}} \\

& 
\textit{Intra} & \textit{Inter} & \textit{Count.} & \textit{Them.} & \textbf{Ave.} & 
\textit{Intra} & \textit{Inter} & \textit{Count.} & \textit{Them.} & \textbf{Ave.} & \\
\midrule

\multicolumn{12}{l}{\textit{\textbf{Open Source Models}}} \\
\hline

Qwen3-VL-235B-A22B 
& 48.5 & 48.5 & 60.4 & 63.7 & 55.1 
& 7.2  & 6.5  & 9.8  & 10.5 & 8.5  
& 34.9 \\                            

\hline
\multicolumn{12}{l}{\textit{\textbf{Proprietary Models}}} \\
\hline

Gemini-2.5-Flash
& 63.5 & 62.8 & 66.4 & 68.1 & 65.2 
& 26.5 & 25.8 & 30.2 & 32.1 & 28.6 
& 46.6 \\                            
\hline

Gemini-2.5-Pro
& \textbf{71.5} & \textbf{70.8} & \textbf{74.6} & \textbf{75.5} & \textbf{73.1} 
& \textbf{33.2} & \textbf{31.5} & \textbf{36.8} & \textbf{39.4} & \textbf{35.2} 
& \textbf{54.2} \\                            

\bottomrule
\end{tabular}
}
\caption{Comparison between Multiple-Choice Reasoning Accuracy and Evidence Score. The \textbf{Overall} score is calculated as $S_{\text{total}} = 0.6 \cdot Acc_{\text{MCQ}} + 0.4 \cdot (Acc_{\text{MCQ}} \cdot S_{\text{evidence}})$, strictly penalizing models that fail to provide correct evidence for their choices. The best scores in each column are highlighted in \textbf{bold}.}
\label{tab:reasoning_evidence_comparison}
\end{table*}

\subsection{Main Results on Video-HolmesV2}
Tab.~\ref{tab:perc_reason_benchmark} presents the comprehensive evaluation of state-of-the-art MLLMs on our Video-HolmesV2 benchmark.

\noindent\textbf{1. The Capability Gap and Proprietary Dominance.} 
There remains a substantial performance gap between proprietary and open-source models. Gemini-2.5-Pro emerges as the state-of-the-art, achieving the highest overall accuracy ($73.2$) and demonstrating robust consistency across all Perception and Reasoning sub-categories. Among the open-source community, the massive Qwen3-VL-235B-A22B establishes the frontier baseline with an overall score of $55.3$, yet it still lags significantly behind the leading closed-source counterpart.  
Interestingly, GPT-4o struggles on our benchmark ($46.9$). We attribute this primarily to its restrictive input constraints (capped at 50 sampled frames). In long-form narratives requiring dense, multi-timestamp audio-visual integration, such extreme sparse sampling inevitably misses fleeting critical evidence and severs long-range causal chains, rendering it ineffective for video reasoning.

\noindent\textbf{2. The Perception-to-Reasoning Degradation.} 
The sub-module breakdown exposes a sharp cognitive cliff in smaller open-source models. For instance, Qwen3-VL-4B achieves a reasonable $47.3$ in foundational Perception but collapses to $27.0$ in higher-order Reasoning. This indicates that while basic spatio-temporal recognition is improving, bridging these observations into multi-hop causal chains over long sequences remains a profound bottleneck.

\noindent\textbf{3. Superior Efficiency of Our Approach.} 
Notably, our framework (\textbf{Ours}, $44.5$) effectively outperforms parameter-heavy models (e.g., Qwen2.5-VL-32B, $43.6$) and omni-models (Qwen2.5-Omni, $40.7$). Despite severe token compression, its strong performance—particularly in \textit{Scene} and \textit{Thematic} reasoning—proves that our Audio-Text Guided strategy successfully preserves critical macro-context and micro-evidence without token explosion.

\noindent\textbf{4. Severe Degradation over Extreme Contexts.} 
Tab.~\ref{tab:video_benchmark} reveals a universal vulnerability: reasoning accuracy degrades significantly as video length increases. The leading Qwen3-VL-235B-A22B drops sharply from $\sim70$ on Short videos to below $45$ on Ultra-long movies, where it is even outperformed by smaller models like Qwen2.5-VL-72B. This proves that simply scaling up parameters cannot resolve "attention dilution" in massive contexts, highlighting the critical necessity of our efficient token compression method (detailed analysis in Appendix~\ref{sec:app_duration_genre}).
\subsection{The Illusion of Accuracy: Why Evidence Matters.} 

Tab.~\ref{tab:reasoning_evidence_comparison} shows Multiple-Choice (MCQ) accuracy with spatio-temporal evidence-aware capabilities (smaller models are excluded due to near-zero evidence scores). A striking paradox emerges with \textbf{Qwen3-VL-235B-A22B}: despite a respectable $55.1$ MCQ accuracy, its Evidence score collapses to $8.5$. This exposes a critical flaw in traditional evaluations: the model's performance is largely an illusion. It relies heavily on parametric memory to guess answers but hallucinates the spatio-temporal proof. 
Conversely, \textbf{Gemini-2.5-Pro} shows healthier reasoning-grounding synchronization (Evidence $35.2$). By strictly computing $S_{\text{total}}$ to penalize ungrounded guessing, our benchmark evaluates \textit{epistemic honesty}—ensuring models genuinely deduce from video facts rather than exploiting statistical priors (More experiments in Appendix~\ref{supp_exper}).
\begin{table*}[t]
    \centering
    \resizebox{\textwidth}{!}{
    \begin{tabular}{l c ccccccc c c c}
        \toprule
        \multirow{2}{*}{Method} & \multirow{2}{*}{Retained Ratio} & \multicolumn{7}{c}{AVUTBench} & VideoMME & ShortVid-Bench & \multirow{2}{*}{Avg. Ratio} \\
        \cmidrule(lr){3-9} \cmidrule(lr){10-10} \cmidrule(lr){11-11}
        & & EL & OR & OM & IE & CC & CM & Avg. & wo & Avg. Score & \\
        \midrule
        \multicolumn{12}{c}{\textit{Qwen2.5-Omni-7B}} \\
        \midrule
        \rowcolor{gray!20}
        Full Tokens & 100\% & 38.2 & 67.8 & 59.6 & 85.6 & 44.1 & 66.7 & 64.5 & 66.0 & 70.5 & 100\% \\
        \midrule
        Random & 40\% & 31.7 & 58.5 & 53.3 & 74.9 & \underline{43.2} & 59.0 & 56.9 & 65.0 & 67.7 & 94.3\% \\
        Random & 55\% & 38.2 & 64.9 & 55.6 & 80.1 & 34.7 & 65.0 & 61.0 & 65.4 & 68.3 & 96.9\% \\
        
        FastV & 35\% & 24.1 & 60.7 & 54.3 & 81.6 & 40.7 & 58.3 & 57.8 & - & 67.9 & 93.8\% \\
        FastV & 50\% & 34.1 & 64.3 & 57.1 & 77.6 & 36.4 & 56.4 & 58.4 & - & 68.0 & 94.3\% \\
        
        DyCoke (V\&A) & 35\% & 32.9 & 62.1 & 54.9 & 74.5 & 39.0 & 58.3 & 57.4 & 65.2 & 68.0 & 94.7\% \\
        DyCoke (V\&A) & 50\% & \textbf{38.8} & 67.2 & \underline{58.2} & 81.9 & 39.0 & 62.4 & 62.0 & 65.5 & 68.5 & 97.5\% \\
        OmniZip & 35\% & 34.1 & \underline{67.5} & 54.6 & 83.7 & 42.4 & 61.2 & 61.0 & 66.1 & 69.0 & 97.6\% \\
        OmniZip & 45\% & \underline{38.4} & 67.2 & 56.9 & \textbf{85.3} & 42.4 & \underline{66.0} & \underline{63.0} & \underline{66.3} & \underline{69.9} & \underline{99.1\%} \\
        \midrule
        \midrule
        Ours & - & 37.8 & \textbf{68.5} & \textbf{61.2} & \underline{85.0} & \textbf{46.3} & \textbf{68.2} & \textbf{65.5} & \textbf{67.8} & \textbf{70.8} & \textbf{101.5\%} \\
        \bottomrule
    \end{tabular}
    }
    \caption{Performance on AVUTBench, VideoMME, and ShortVid-Bench.}
    \label{tab:datasets_performance}
\end{table*}
\begin{table*}[t]
    \centering
    \resizebox{\textwidth}{!}{
    \begin{tabular}{l c ccccccccc}
        \toprule
        Method & Retained Ratio & \begin{tabular}[c]{@{}c@{}}Tech \&\\Science\end{tabular} & \begin{tabular}[c]{@{}c@{}}Culture \&\\Politics\end{tabular} & \begin{tabular}[c]{@{}c@{}}Daily\\Life\end{tabular} & \begin{tabular}[c]{@{}c@{}}Film \&\\TV\end{tabular} & Perform. & Games & Sports & Music & Avg. \\
        \midrule
        \multicolumn{11}{c}{\textit{Qwen2.5-Omni-7B}} \\
        \midrule
        \rowcolor{gray!20}
        Full Tokens & 100\% & 52.4 & 50.1 & 48.5 & 44.6 & 43.8 & 41.6 & 41.6 & 47.3 & 46.8 \\
        \midrule
        Random & 55\% & 47.1 & 47.0 & 44.4 & 41.2 & 40.0 & 40.1 & 40.1 & 46.3 & 43.6 \\
        FastV & 50\% & \underline{48.8} & 47.4 & 44.2 & \underline{44.1} & \underline{41.2} & 38.3 & 40.0 & \underline{46.6} & 44.3 \\
        DyCoke (V\&A) & 50\% & 48.4 & \underline{49.9} & 46.7 & 41.4 & 39.9 & \underline{40.8} & 40.2 & 46.5 & 44.6 \\
        OmniZip & 35\% & 48.3 & 49.5 & \underline{47.6} & 42.5 & 40.1 & 40.2 & \underline{42.3} & 46.3 & \underline{45.3} \\
        \midrule
        \midrule
        Ours & - & \textbf{51.5} & \textbf{50.5} & \textbf{48.8} & \textbf{45.1} & \textbf{42.5} & \textbf{41.9} & \textbf{42.8} & \textbf{47.6} & \textbf{47.2} \\
        \bottomrule
    \end{tabular}
    }
    \caption{Detailed performance across different categories on WorldScene.}
    \label{tab:worldscene_performance}
    \vspace{-15pt}
\end{table*}
\subsection{Results on Other Benchmarks.}
As shown in Tab.~\ref{tab:datasets_performance}, our method using {Qwen2.5-Omni-7B} not only outperforms recent compression baselines but remarkably surpasses the uncompressed \textit{Full Tokens} baseline ($101.5\%$ relative performance). This proves our bimodal selection acts as a powerful noise filter to prevent attention dilution. Crucially, unlike ratio-based methods that still suffer from token explosion on long videos, our framework operates on a \textbf{fixed token budget}. While matching OmniZip's $\sim40\%$ retention under a standard 128-frame setting, our fixed-budget design uniquely scales to massive frame counts without memory bottlenecks.
This robust superiority generalizes across diverse genres (Tab.~\ref{tab:worldscene_performance}). Our approach achieves SOTA scores across all eight WorldScene categories, yielding a $47.2$ average that again beats the Full Tokens baseline ($46.8$). This confirms our dual-stream scoring seamlessly adapts to varying narrative paces, enhancing global understanding despite extreme token reduction.

\begin{table*}[t]
    \centering
    \begin{minipage}[t]{0.4\textwidth}
        \centering
        \resizebox{\linewidth}{!}{
        \begin{tabular}{c c c c c}
            \toprule
            \textbf{TGC} & \textbf{AGC} & \textbf{DSFDC} & \textbf{VideoMME} & \textbf{WorldScene} \\
            \midrule
            \ding{55} & \ding{55} & \ding{55} & 66.0 & 46.8 \\
            \checkmark & \ding{55} & \ding{55} & 67.0 & 46.5 \\
            \ding{55} & \checkmark & \ding{55} & 63.3 & 45.1 \\
            \checkmark & \checkmark & \ding{55} & 67.4 & 46.9 \\
            \checkmark & \checkmark & \checkmark & \textbf{67.8} & \textbf{47.2} \\
            \bottomrule
        \end{tabular}
        }
        \caption{Ablation study of components (Text-Guided, Audio-Guided, Dual-Stream Compress).}
        \label{tab:module_ablation}
    \end{minipage}%
    \hfill 
    \begin{minipage}[t]{0.58\textwidth}
        \centering
        
        \resizebox{\linewidth}{!}{
        \begin{tabular}{l ccccc c}
            \toprule
            \multirow{2}{*}{\textbf{Settings}} & \multicolumn{6}{c}{\textbf{Token Combinations}} \\
            \cmidrule(lr){2-7}
            & Comb 1 & Comb 2 & Comb 3 & \cellcolor{gray!20}\textbf{Ours (Best)} & Comb 5 & Full \\
            \midrule
            Visual Tokens & 4096 & 4096 & 8192 & \cellcolor{gray!20}\textbf{8192} & 12288 & - \\
            Audio Tokens  & 1024 & 2048 & 1024 & \cellcolor{gray!20}\textbf{2048} & 4096 & - \\
            \midrule
            VideoMME      & 66.5 & 66.7 & 67.4 & \cellcolor{gray!20}\textbf{67.8} & \underline{67.6} & 66.0 \\
            WorldScene    & 45.7 & 46.2 & 46.9 & \cellcolor{gray!20}\textbf{47.2} & \underline{47.3} & 46.8 \\
            \bottomrule
        \end{tabular}
        }
        \caption{Ablation on Token combinations. The 8192/2048 setting achieves the optimal balance.}
        \label{tab:token_ablation}
    \end{minipage}
\vspace{-20pt}
\end{table*}
\subsection{Ablation Study}

\noindent\textbf{Effectiveness of Key Components.} 
Tab.~\ref{tab:module_ablation} isolates module contributions. Using Audio-Guided Compression (AGC) alone degrades performance by discarding silent yet critical visual frames. However, fusing it with Text-Guided Compression (TGC) boosts VideoMME from $67.0$ to $67.4$. Finally, adding Dual-Stream Fusion and Differentiated Compression (DSFDC) achieves the optimal $67.8$, proving that deduplicating core tokens while preserving background context massively outshines naive hard-pruning.

\noindent\textbf{Optimal Token Allocation.} 
Tab.~\ref{tab:token_ablation} examines token budgets. Accuracy steadily improves as budgets scale up to our default 8192 visual and 2048 audio tokens. Crucially, expanding further (12288 visual / 4096 audio) actually causes a slight performance drop. This empirically validates the "attention dilution" issue: excessive tokens reintroduce background noise and distract reasoning. Thus, the 8192/2048 setting strikes the balance between reasoning accuracy and computational efficiency.

\section{Conclusion}
We introduce a novel audio-visual benchmark for extreme-length videos, alongside a Spatio-Temporal Evidence-Aware Evaluation protocol to rigorously assess fine-grained multimodal reasoning. We observe that massive token counts in long videos severely overwhelm MLLM context limits, causing attention dilution and the loss of critical details. To address this, we propose a Audio-Text Guided Token Compression framework. By prioritizing the retention of crucial audio-visual tokens within a fixed budget, our method effectively filters redundancy, allowing models to remarkably surpass uncompressed full-token baselines.

\clearpage  


%
%
\bibliographystyle{splncs04}
\bibliography{main}

\clearpage
\appendix

\appendix
\begin{center}{\bf \Large Appendix}\end{center}
\section{Details of Initial Video Collection and Technical Curation}
\label{sec:app_video_collection}

In Section 3.1 of the main text, we briefly outlined our video selection and MLLM-based evaluation pipeline. This section provides the comprehensive methodology, elaborating on the specific filtering criteria for various video formats and detailing the exact definitions of the five core dimensions utilized in our Multi-Dimensional Analyzability Assessment.

Our data construction begins with a rigorous selection process designed to curate a high-quality, contamination-free dataset tailored for audio-visual reasoning. We prioritize videos from award-winning channels (e.g., \textit{Omeleto}, Oscar-nominated shorts) that feature professional cinematography and sound design, enforcing a minimum resolution of $1080$p and the availability of clear audio tracks with subtitles to ensure precise multimodal grounding. To mitigate data contamination and evaluate genuine zero-shot capabilities, we apply a strict temporal cutoff, excluding content released prior to 2024 that may have been ingested by current foundation models.

We further filter the candidate pool (about 3704 videos) based on narrative density and structural diversity. The videos are stratified into distinct duration intervals that correspond to specific content formats: \textbf{short films (10--30 min)} which test rapid narrative development, \textbf{episodic TV series and documentaries (30--60 min)} focusing on evolving character arcs, and \textbf{feature-length movies (60--150 min)} that require maintaining long-range memory over complex storylines. We also impose an engagement threshold of over $500$ user comments as a proxy for interpretability. Finally, to guarantee multimodal interdependence, we systematically exclude categories dominated by a single modality or lacking causal logic, such as news reports (audio-dominant), sports matches (rule-based), and tutorials. This filtering yields a collection of visually and acoustically rich narratives, serving as a robust foundation for the subsequent analyzability assessment.

\textbf{Multi-Dimensional Analyzability Assessment.}
Moving beyond traditional text-based evaluation, we employ a consortium of state-of-the-art native Multi-modal MLLMs (e.g., Gemini 3 Pro, GPT-5.2, Qwen3-VL-235B-thinking) to directly assess the reasoning potential of each video. For ultra-long sequences exceeding context limits, we adopt a hierarchical ``segment-caption-aggregate'' strategy to preserve global context. To rigorously quantify the suitability of a video for our benchmark, we define five core dimensions, scored on a scale from 1 to 10:\\
\textbf{1. Audio-Visual Complementarity:} Evaluates the degree of interdependence between visual and auditory channels. A high score indicates that neither modality is sufficient in isolation; correct comprehension requires synthesizing information from both (distinguishing a sarcastic tone from a serious facial expression).\\
\textbf{2. Cross-Modal Causal Complexity:} Measures the depth of cause-and-effect chains that span across modalities and time. High-scoring videos feature intricate logic where an event in one modality (a specific sound cue) precipitates a consequence in another (a visual reaction) at a later timestamp, requiring multi-hop deduction.\\
\textbf{3. Dynamic Character Arc:} Assesses whether key entities or their relationships undergo meaningful transformations over time. Crucially, this dimension requires that such evolution be evidenced through both visual changes (e.g., appearance, distance) and auditory shifts (e.g., tone, vocabulary), ensuring the model must track long-term state dynamics.\\
\textbf{4. Narrative Coherence:} Gauges the structural integrity and logical consistency of the storyline. A high score guarantees a well-structured plot with clear motivations and resolutions, filtering out videos with disjointed editing or internal logical fallacies that would render reasoning ambiguous.\\
\textbf{5. Audience Critical Reception:} Serves as a meta-metric for narrative depth. Beyond simple popularity, this dimension analyzes the sentiment and content of user comments to determine if the video provokes high-quality intellectual discussion (debates on plot details, theories about endings) rather than mere superficial reactions or negativity.

\subsection{Extended Details of the Annotation Pipeline}
\label{sec:app_annotation}

In Section~\ref{sec:annotation_pipeline} and Fig~\ref{fig:pipeline} of the main text, we presented an overview of our four-phase data annotation and quality control pipeline. This section provides the comprehensive technical details, including specific model deployments, prompt engineering strategies, and algorithmic configurations for each phase. 

We propose a unified, scalable annotation pipeline that transcends traditional clip-based captioning by leveraging the native long-context capabilities of state-of-the-art MLLMs (e.g., {Gemini 3 Pro}). Unlike previous approaches that fragment narrative flow, our method processes entire video streams to capture long-range dependencies. The pipeline is structured into three rigorous phases: Stratified Generation, Unimodal-Blind Filtration, and Multi-Model Cross-Verification.

\paragraph{Phase 1: Stratified Question Generation.}
Recognizing that narrative complexity varies significantly across formats, we employ a differentiated generation strategy to maximize the depth of reasoning.

\textbf{1. Direct Long-Context Prompting (Short Films, 10--30 min).} For concise narratives, we leverage the full-context window of {Gemini 3 Pro}. The model ingests the complete video stream alongside time-stamped subtitles. To prevent generic QA generation, we employ a specific \textit{Constraint-Based Prompting} strategy. The model is explicitly instructed to focus on ``Audio-Visual Dissonance'' (where audio contradicts vision) and ``Cross-Modal Synthesis'' (where visual details explain auditory cues). The output includes not just the QA pair, but structured \textit{spatio-temporal evidence}, pointing to precise visual and audio timestamps that are explicit grounded in the video timeline.

\textbf{2. Context-Primed Interactive Refinement (Episodic Series, 30--60 min).} TV dramas often rely on implicit lore or pre-existing character dynamics that a standalone video file cannot convey. To address this, we implement an \textit{Expert-in-the-Loop Contextual Priming} protocol. Before generation, domain experts inject a ``Series Summary''—a structured summary of character relationships and prior plot points—into the {Gemini 3 Pro} session context. This priming enables the model to generate deep, motivation-centric questions rather than surface-level observation queries. The generation process is interactive: experts review initial seeds and dynamically adjust the prompt focus (``Focus on the evolution of the relationship between Entity A and B'') via Google AI Studio, ensuring the questions align with the dramatic arc.

\textbf{3. Comment-Driven Inverse Reasoning (Feature Films, $>$60 min).} For complex movies, we harness collective human intelligence to mine high-difficulty reasoning chains. We scrape the top 250 user comments per video and employ a dual-model scoring system using GPT-5.2-High and Qwen3-235B-A22B-Thinking. Each comment is rigorously scored from 1 to 10 based on its logical depth and relevance to the plot. We retain only the \textit{Semantic Intersection} of comments rated $\ge 8$ by both models. These high-quality insights serve as seed prompts for {Gemini 3 Pro}, which is tasked with \textit{Inverse Reasoning}: reverse-engineering the user's insight into a multiple-choice question and then re-scanning the full video to locate the concrete audio-visual evidence that supports this high-level interpretation.

\paragraph{Phase 2: The ``Unimodal Trap'' Blind Filtration.}
To strictly enforce the requirement for multimodal reasoning, we introduce an \textit{Unimodal Trap} mechanism using {gpt-4o-mini} as an adversarial filter. For every generated QA pair, we create two ablated baselines: a \textit{Visual-Only} input (sampled frames at 1fps without audio) and an \textit{Audio-Only} input (subtitles without video). These are fed to {gpt-4o-mini}. If this lightweight model can answer the question with high confidence using either single modality, the question is deemed a ``pseudo-multimodal'' sample and is automatically discarded. This ensures that our benchmark exclusively consists of questions where the joint probability $P(A|V, Audio)$ is significantly higher than any unimodal marginal probability.

\paragraph{Phase 3: Segment-Level Cross-Verification.}
While generation utilizes global context, verification requires the precision of reasoning-heavy models which may have shorter context windows or higher costs. We leverage the \textit{Semantic Event Segmentation} algorithm (detailed in Appendix~\ref{sec:segmentation}) to divide the video into coherent, non-overlapping narrative units (max 10 minutes). This segmentation allows us to employ a consortium of diverse, high-performance validators: {Gemini 3 Pro} , {Gemini 3 Flash (Thinking)}, {GPT-5.2-High}, and {GPT-o3}. For each QA pair, the specific event segment containing the ground-truth evidence is isolated and fed to this ensemble. A sample is validated only if a majority of the ensemble converges on the same answer and retrieves evidence timestamps that align with the generation phase (Intersection over Union $> 0.75$).

\paragraph{Phase 4: Expert Adjudication for Hard Negatives.}
Samples where the validator ensemble disagrees ($D_{\text{video}} > \tau_{div}$) or fails to reproduce the answer are flagged as \textit{Hard Negatives} and routed to a dedicated Double-Blind Expert Review queue. Two human specialists independently assess these cases to distinguish between model hallucination and genuine reasoning difficulty. Experts evaluate the validity of the reasoning chain and the existence of objective audio-visual evidence. Only cases where both experts confirm the logic is deterministic but requires complex integration are retained. These validated samples constitute the \textit{Hard Subset} of our benchmark, representing the upper bound of current AI reasoning capabilities. More details in Appendix~\ref{sec:expert_protocol}

\subsection{Semantic Event Segmentation Pipeline}
\label{sec:segmentation}

To overcome the limitations of arbitrary temporal splitting (fixed-duration clips), which often disrupts narrative continuity, we devised a hierarchical segmentation pipeline. This approach synergizes bottom-up visual discontinuity detection with top-down semantic understanding to partition long-form videos into coherent, meaningful units suitable for precise verification.

\paragraph{Global Visual Coordinate System.}
The foundation of our pipeline is the establishment of a discrete visual timeline. We first process the raw video stream using {AutoShot}, a standard boundary detection tool, to generate a \textbf{Global Shot Map}. This map acts as a rigorous coordinate system, indexing every shot $s_i$ with precise start and end timestamps. By grounding all subsequent operations in this shot map, we ensure that event boundaries always align with natural visual transitions (cuts/fades), preventing the jarring artifacts that occur when splitting videos mid-shot.

\paragraph{Sliding-Window Semantic Identification.}
We employ an overlapping sliding window strategy (window size $= 10$ min, stride $= 5$ min) to traverse the video. For each window, the video content and its corresponding sequence of shot indices are fed into a high-efficiency VLM ( {Gemini 3 Flash Thinking}).
A specialized prompt instructs the model to identify the single most significant narrative event contained within the window. Crucially, the model is constrained to output the boundaries of this event as \textbf{Start and End Shot IDs} rather than fuzzy timestamps. This structural constraint forces the VLM to explicitly align its semantic interpretation with the pre-calculated visual structure.

\paragraph{Temporal Consolidation and Output.}
Since narrative events often span across the artificial boundaries of sliding windows, we apply a merging algorithm to consolidate the local predictions. Event proposals from adjacent windows that exhibit significant overlap in their shot ID sequences (Intersection over Union $> 0.7$) are unified into a single continuous segment. After pruning anomalous fragments (segments $< 30$ seconds), we obtain the final list of \textbf{Event Segments}. Each segment object encapsulates the semantic title, global timestamps, and a \textbf{local shot manifest}. This structured representation provides the downstream annotation and verification models with both the macro-narrative context and the micro-visual granularity required for identifying precise audio-visual evidence.

\subsection{Unified Expert Adjudication Protocol}
\label{sec:expert_protocol}

To ensure the integrity of our benchmark, we establish a rigorous, two-stage human-in-the-loop adjudication protocol. This unified framework addresses both the selection of controversial source videos (Stage I) and the validation of complex QA pairs (Stage II), utilizing human expertise to systematically distinguish between \textit{undesirable ambiguity} (noise) and \textit{meaningful complexity} (hardness).

\paragraph{Expert Qualification and Training.}
We recruited a specialized team of domain experts, primarily consisting of graduate researchers with backgrounds in media analysis, cognitive science, and linguistics. Unlike crowdsourced annotators, these experts possess the analytical skills required to deconstruct long-form narrative structures. Before entering the adjudication pool, all candidates underwent a rigorous onboarding phase, evaluating a standard set of 50 samples. Only those achieving an agreement rate exceeding 90\% against ground truth were qualified, ensuring a baseline of high-consistency judgment.

\paragraph{Stage I: Source Video Adjudication (Curation Phase).}
The first layer of human intervention targets source videos exhibiting high model divergence ($D_{\text{video}} > 1.2$) during the quantitative curation process. Experts assess whether the disagreement stems from poor video quality or high reasoning difficulty based on two criteria:
\begin{itemize}
    \item \textbf{Logical Determinism:} Verifying that the narrative, while potentially intricate or non-linear, possesses a coherent internal logic with a deterministic resolution, effectively filtering out videos with plot holes or artistic abstractness that precludes objective reasoning.
    \item \textbf{Audio-Visual Indispensability:} Confirming that understanding the narrative strictly requires the integration of both visual and auditory signals. Videos where the plot is understandable via dialogue alone (radio-play style) or visuals alone (silent-film style) are rejected to enforce multimodal rigor.
\end{itemize}

\paragraph{Stage II: Reasoning Chain Verification (Annotation Phase).}
The second layer targets QA pairs flagged as ``Hard Negatives''—instances where the automated validator ensemble (GPT-5.2, Gemini 3 pro) disagreed on the answer or failed to find evidence. Experts validate these pairs to salvage high-difficulty questions:
\begin{itemize}
    \item \textbf{Causal Validity:} Experts verify that the reasoning chain linking the evidence to the answer is logically sound and free from ``leaps of logic,'' ensuring the difficulty arises from the depth of inference rather than question phrasing.
    \item \textbf{Precise Grounding:} Experts rigorously check the generated \textit{spatio-temporal evidence}. They must confirm that the specific visual frames and audio cues cited by the model are objectively present and sufficient to support the answer.
\end{itemize}

\paragraph{Consensus Mechanism and Hard Subset.}
We employ a strict \textbf{Double-Blind Review} protocol for both stages. Two experts independently assess each flagged sample. We monitor Inter-Annotator Agreement (IAA) using Cohen’s Kappa ($\kappa$). In cases of disagreement ($\kappa < 0.8$), a senior author acts as the final arbitrator.
Samples that survive this gauntlet—videos that are complex but logical, and QA pairs that are difficult but verifiable—form the \textbf{Hard Subset} of our benchmark. This subset represents the upper bound of current machine reasoning capabilities, offering a stress test free from label noise.

\subsection{Validation of the Multi-Modal Evidence Evaluation}
\label{sec:appendix_judge_validation}

Traditional N-gram matching metrics (e.g., BLEU, ROUGE) consistently fall short when assessing the nuanced logic of long-form, spatio-temporal reasoning, as they fail to capture deeply coupled cross-modal dependencies. To circumvent these limitations without compromising evaluation scalability, we operationalize Qwen3-VL-235B-A22B-Thinking as our automated judge. Endowed with advanced Chain-of-Thought (CoT) capabilities, this evaluator is tasked with meticulously cross-verifying lengthy multimodal context windows. To empirically validate its reliability and insulate our benchmark against the vulnerabilities of the ``LLM-as-a-judge'' paradigm, we orchestrated a rigorous double-blind study. Specifically, 300 reasoning paths—sampled across diverse task categories and generated by various models—were independently scored by human experts who remained entirely agnostic to the source models' identities.

\noindent\textbf{Deterministic Evaluation Rubric.} 
To anchor the evaluation mathematically, both human annotators and the automated judge strictly adhere to a three-tier deterministic rubric. A full score ($1.0$) is contingent upon flawless multimodal alignment, demanding that the model explicitly and accurately captures both visual (V) and acoustic (A) Ground Truth clues without logical fracture. Conversely, partial credit ($0.6$) is assigned when a correct conclusion is reached but the supporting evidence remains unimodal or structurally incomplete (citing dialogue while neglecting crucial visual corroboration). Outputs devolving into factual hallucinations, temporal misalignments, or those completely untethered from valid bimodal timestamps receive a null score ($0.0$).

\noindent\textbf{Statistical Validation of Judge Competence.} 
Rather than relying solely on absolute agreement rates—which can be artificially inflated by trivial queries—we substantiate the judge's efficacy through a multi-dimensional statistical lens, as summarized in Tab.~\ref{tab:human_machine_agreement}. The Pearson correlation coefficient ($r$) is utilized to capture the linear trajectory of scoring trends, while Cohen’s Kappa ($\kappa$) acts as a stringent penalty against coincidental consensus. Remarkably, the automated evaluator achieves a robust overall absolute agreement of 90.2\%. More importantly, the Cohen's Kappa scores consistently exceed the 0.80 threshold across all reasoning axes. Within statistical literature, a $\kappa > 0.80$ is unequivocally recognized as ``almost perfect'' agreement, thereby affirming that our Qwen3-235B-driven pipeline consistently mirrors expert human judgment, even in the highly demanding domain of counterfactual reasoning.
\begin{table}[htbp]
\centering
\resizebox{0.95\columnwidth}{!}{
\begin{tabular}{lccc}
\toprule
\textbf{Task Category} & \textbf{Absolute Agreement (\%)} & \textbf{Pearson ($r$)} & \textbf{Cohen's $\kappa$} \\
\midrule
Intra-Event Reasoning    & 92.4 & 0.88 & 0.85 \\
Inter-Event Causal       & 89.6 & 0.84 & 0.81 \\
Global \& Thematic       & 88.2 & 0.85 & 0.82 \\
Counterfactual Reasoning & 90.8 & 0.87 & 0.84 \\
\midrule
\rowcolor{gray!10} \textbf{Overall (500 samples)} & \textbf{90.2} & \textbf{0.86} & \textbf{0.83} \\
\bottomrule
\end{tabular}
}
\caption{\textbf{Human-Machine Consensus across Reasoning Tasks.} Evaluated on 300 blind samples, the Qwen3-based judge achieves ``almost perfect'' true agreement with human experts (indicated by Cohen's $\kappa > 0.80$), successfully ruling out coincidental scoring inflations.}
\label{tab:human_machine_agreement}
\end{table}

\begin{figure*}[htbp]
    \centering
    \includegraphics[width=1.0\textwidth]{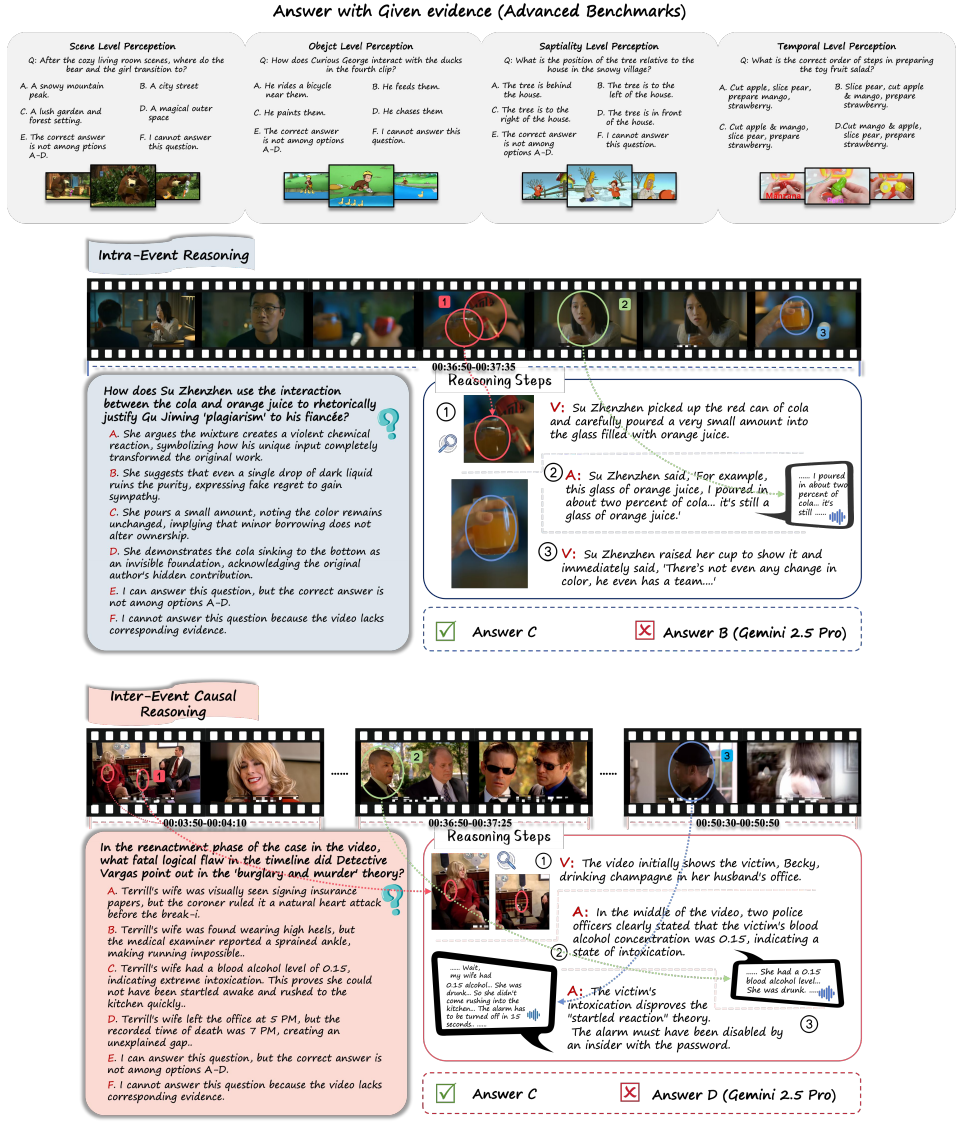}
    \caption{\textbf{Foundational Perception and Intra/Inter-Event Reasoning.} The top section demonstrates standard perception tasks. The middle and bottom sections showcase complex reasoning where Ground Truth (GT) visual (V) bounding boxes and acoustic (A) dialogues are inextricably linked. Gemini 2.5 Pro's failure in these instances highlights its inability to maintain fine-grained cross-modal alignment and bridge long-range temporal gaps.}
    \label{fig:app_perception}
\end{figure*}

\begin{figure*}[htbp]
    \centering
    \includegraphics[width=1.0\textwidth]{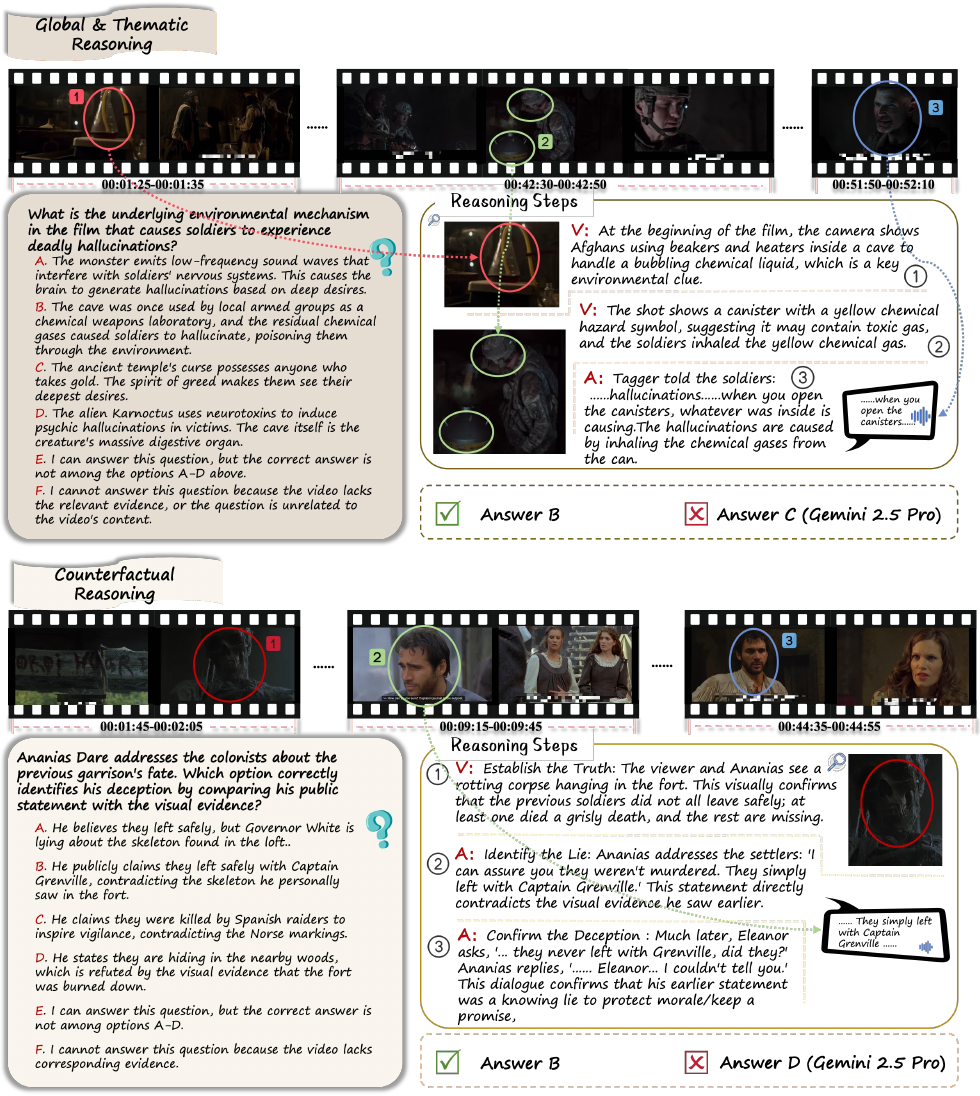}
    \caption{\textbf{Global Synthesis and Counterfactual Reasoning.} To deduce environmental mechanisms or detect character deception, models must synthesize globally scattered spatio-temporal evidence and resolve direct audio-visual contradictions (visual truth vs. deceptive dialogue). The depicted GT evidence chains expose the limitations of current MLLMs in extended multimodal reasoning.}
    \label{fig:app_reasoning}
\end{figure*}

\section{Qualitative Analysis and Evidence-Aware Reasoning}
\label{sec:appendix_qualitative}

To comprehensively elucidate the rigorous demands of Video-HolmesV2, we present a detailed qualitative analysis of our proposed taxonomy in Figure~\ref{fig:app_perception} and Figure~\ref{fig:app_reasoning}. While foundational perception tasks (Figure~\ref{fig:app_perception}, top) evaluate a model's basic capability to localize explicit spatial-temporal elements—such as scenes, objects, spatial relations, and temporal sequences—the true challenge of our benchmark lies in the advanced reasoning categories. To solve these complex queries, models cannot rely on parametric priors or unimodal shortcuts; they must formulate an unbroken chain of Ground Truth (GT) spatio-temporal audio-visual evidence. In the illustrated examples, we explicitly trace this GT evidence—visualized as specific timestamped bounding boxes (V) and acoustic transcripts (A)—to demonstrate how inextricably linked multimodal clues govern the final logical deduction. Notably, we intentionally highlight failure cases from a leading state-of-the-art model, Gemini 2.5 Pro, exposing its vulnerability in maintaining deep audio-visual coupling over extended contexts.

\noindent\textbf{Intra-Event Metaphorical Reasoning.} 
As depicted in the middle of Figure~\ref{fig:app_perception}, intra-event reasoning challenges models to align fine-grained visual actions with concurrent acoustic metaphors within a localized temporal window. To understand how the character rhetorically justifies ``plagiarism'', the model must precisely ground the visual action of pouring a minuscule amount of cola (V1) into orange juice, map it to the dialogue confirming ``two percent of cola... it's still orange juice'' (A2), and observe the unchanged visual color of the mixture (V3). Gemini 2.5 Pro fails this task (selecting B) because it misses the critical visual evidence of the unchanged color, highlighting a failure to establish a micro-level spatio-temporal consensus between the physical action and the dialogue's metaphorical intent.

\noindent\textbf{Inter-Event Causal Bridging over Long Durations.} 
The bottom of Figure~\ref{fig:app_perception} exemplifies the necessity of long-range temporal memory. The query requires identifying a fatal logical flaw in a detective's reconstructed timeline. The GT evidence chain spans nearly an hour of video: it begins with an early visual confirmation of the victim drinking champagne (V1), which must be cross-verified against a much later acoustic police report detailing a 0.15 blood alcohol concentration (A2). This audio-visual synthesis conclusively disproves the ``startled awake'' theory (A3). Gemini 2.5 Pro outputs an incorrect assumption (Answer D) due to its inability to bridge the massive temporal gap. It loses the visual state (intoxication) established early in the narrative when processing the later acoustic medical report, demonstrating the severe context truncation prevalent in current MLLMs.

\noindent\textbf{Global and Thematic Synthesis.} 
Moving to macroscopic reasoning, the top of Figure~\ref{fig:app_reasoning} requires the extraction of underlying environmental mechanisms. The GT evidence is heavily fragmented across the entire video: early visual clues of Afghans handling chemicals (V1), subsequent visual recognition of yellow chemical hazard canisters (V2), and a decisive, temporally distant acoustic explanation that inhaling the gas causes hallucinations (A3). Answering correctly demands an omni-modal synthesis of these scattered spatio-temporal anchors to deduce the overarching environmental rule. Gemini 2.5 Pro fails (Answer C) by falling back on narrative clichés (``ancient curses''), proving that without a rigorous mechanism to retain key evidence tokens over long durations, models default to hallucinated priors rather than extracting factual, globally dispersed clues.

\noindent\textbf{Counterfactual and Deception Detection.} 
Finally, the bottom of Figure~\ref{fig:app_reasoning} illustrates one of the most challenging paradigms: resolving direct contradictions between modalities. To identify a character's public deception, the model must first establish the objective truth via a localized visual anchor—a rotting corpse hanging in the fort (V1). It must then contrast this undeniable visual fact against the character's deceptive public speech claiming the colonists ``left safely'' (A2), further confirmed by a later private acoustic confession (A3). Gemini 2.5 Pro's failure (Answer D) underscores a critical flaw in current visual-centric architectures: when confronted with conflicting audio-visual signals, they often fail to prioritize grounded visual reality, instead being misled by explicit but deceptive textual/auditory dialogue. This highlights the indispensable need for explicit audio-visual cross-verification in long-form video understanding.

\section{Details of Method}
\subsection{Details on Text-Guided Visual Scoring}
\label{text_scoring_details}
\begin{figure*}[t] 
  \centering
  \includegraphics[width=0.95\textwidth]{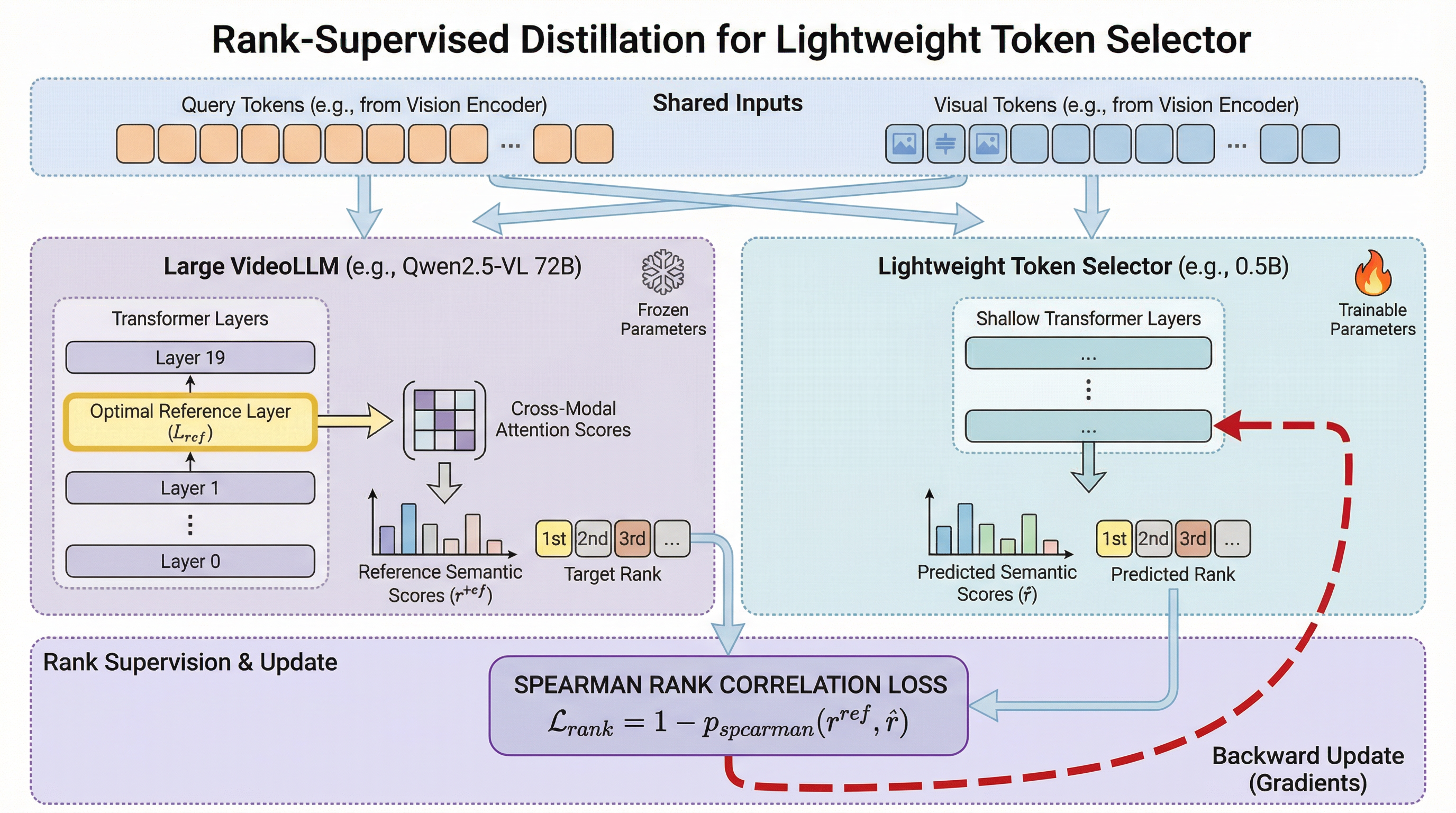}
  \caption{\textbf{Rank-Supervised Distillation for the Lightweight Token Selector.} 
  To circumvent the prohibitive computational cost of massive encoders during inference, we distill the semantic attention patterns of a large frozen teacher (Qwen2.5-VL 72B) into a lightweight student selector (0.5B). 
  Instead of regressing exact attention values, the student is trained to predict the relative importance ranking of visual tokens via a differentiable \textbf{Spearman Rank Correlation Loss}. 
  This ensures the lightweight model inherits the teacher's ``needle-in-a-haystack'' retrieval capabilities while maintaining high inference efficiency.}
  \label{fig:distillation}
\end{figure*}
In Section \ref{sec:text_guided_scoring} of the main text, we introduced a lightweight text-guided token selector. Here, we provide a comprehensive elucidation of the teacher signal extraction and the differentiable rank-supervised distillation process.

\textbf{Reference Layer Selection in Teacher Model.} 
Large VideoLLMs process tokens through dozens of transformer layers. Empirical observations reveal a "semantic alignment trajectory": shallow layers focus on low-level visual features, extremely deep layers suffer from attention collapse due to autoregressive generation, while intermediate layers capture the most faithful cross-modal semantic correlations. We systematically evaluate the retrieval accuracy of different layers using a "needle-in-a-haystack" visual probing task and empirically designate layer $L_{ref}$ (the 19th layer in a 72B model) as the reference layer. The attention score $\mathbf{r}^{ref}$ extracted from this layer serves as the optimal pseudo-label for semantic relevance.

\textbf{Differentiable Spearman Rank Distillation.} 
The objective of our token selector (0.5B parameters) is not to regress the exact attention values of the 72B model—which are highly scale-dependent and volatile—but to replicate its \textit{preference ranking} over the visual tokens. To this end, we employ the Spearman Rank Correlation Coefficient. 

Let $\mathbf{r}^{ref} \in \mathbb{R}^N$ be the teacher's scores and $\hat{\mathbf{r}} \in \mathbb{R}^N$ be the student's predictions. The exact rank operation $\text{argsort}(\cdot)$ is non-differentiable, making standard backpropagation infeasible. To address this, we adopt a Fast Differentiable Sorting algorithm to compute the soft rank $R(\cdot)$. The Spearman rank correlation $\rho_{spearman}$ is mathematically computed as the Pearson correlation between the rank variables:
\begin{equation}
    \rho_{spearman}(\mathbf{r}^{ref}, \hat{\mathbf{r}}) = \frac{\sum_{i=1}^{N} (R(r_i^{ref}) - \frac{N+1}{2}) (R(\hat{r}_i) - \frac{N+1}{2})}{\sqrt{\sum_{i=1}^{N} (R(r_i^{ref}) - \frac{N+1}{2})^2 \sum_{i=1}^{N} (R(\hat{r}_i) - \frac{N+1}{2})^2}}
\end{equation}
where we utilize the property that the mean rank $\bar{R} = \frac{N+1}{2}$. The final loss function is defined as $\mathcal{L}_{rank} = 1 - \rho_{spearman}(\mathbf{r}^{ref}, \hat{\mathbf{r}})$. Minimizing this loss strictly penalizes the lightweight selector when it assigns a high score to a visual token that the teacher model deemed irrelevant to the text query.

\subsection{Details on Audio-Guided Visual Scoring}
\label{audio_scoring_details}

In Section \ref{sec:audio_guided_scoring}, we described the extraction of salient audio anchors and their subsequent utilization for scoring visual tokens. This section elaborates on the mathematical formulations and the rationale behind our design choices.

\textbf{Intra-Modal Audio Attention Matrix.} 
Modern audio encoders (Whisper-based or WavLM-based architectures) internally model the temporal dependencies of audio frames. The self-attention matrix $\mathbf{M}$ used for our audio saliency estimation is computed directly from the query ($\mathbf{Q}_a$) and key ($\mathbf{K}_a$) projections of the audio encoder's final layer:
\begin{equation}
    \mathbf{M} = \text{Softmax}\left(\frac{\mathbf{Q}_a \mathbf{K}_a^T}{\sqrt{d}}\right)
\end{equation}
Tokens corresponding to sudden energy bursts or distinct phonetic features naturally act as "sinks" for attention probability mass, resulting in higher column sums in $\mathbf{M}$. Filtering the initial sequence to the top-$K_a$ tokens (where $K_a \ll N_a$) is an essential denoising step; it prevents pervasive, low-energy ambient sounds from spuriously inflating the cross-modal similarity of static visual backgrounds in the subsequent stage.

\textbf{Rationale for Max-Pooling in Cross-Modal Similarity.} 
When computing the audio-guided visual score $S_{audio}^{(i)}$ for a visual token $v_i$, we calculate its cosine similarity against all $K_a$ retained audio anchors and take the maximum value ($\max$), rather than the average ($\text{mean}$). The physical intuition behind this design is that an objective visual action (a glass shattering) typically corresponds to a \textit{singular, highly specific} auditory event rather than an aggregation of all sounds present in the environment. 


A mean-pooling strategy would erroneously dilute the visual score of the shattering glass if the audio anchor set simultaneously contained unrelated background music tokens. The max-pooling operation, formulated as
\[
S_{\text{audio}}^{(i)}
=
\max_{j}
\left(
\operatorname{cos\_sim}
\left(
v_i, \hat{a}_j
\right)
\right),
\]
ensures that a visual token receives a high score as long as it strongly resonates with \textit{at least one} distinct acoustic anchor, perfectly preserving the discrete nature of multimodal events.

\subsection{Details on Differentiated Token Compression}
\label{compression_details}

In Section \ref{sec:fusion_and_compression}, we introduced the Dual-Stream Fusion and Differentiated Compression mechanism. Here, we detail the computational complexity optimizations and the hyperparameter configurations.

\textbf{Complexity Reduction via Masked Adjacency.} 
A naive implementation of token merging requires computing a global $\mathcal{O}(N^2)$ pairwise similarity matrix, which is computationally catastrophic for long videos where $N > 50,000$. Our framework radically bypasses this bottleneck via the binary mask $\mathcal{M}$. By strictly constraining the similarity computation to \textit{local spatio-temporal neighborhoods} (matching a token at $(t, x, y)$ only with its temporal neighbors $(t\pm1, x, y)$ or spatial adjacent windows) and ensuring both tokens satisfy $\mathcal{M}=1$, the complexity is reduced to $\mathcal{O}(N \cdot w)$, where $w$ is a small, constant window size. This localized, mask-guided matching preserves the physical geometry of the video and operates efficiently on the GPU.

\textbf{Hyperparameter Selection and Similarity Thresholds.} 
The efficacy of our dynamic quota allocation hinges on two crucial hyperparameters: the intra-set similarity threshold $\tau_{intra}$ and the background scavenging ratio $M$.

\textbf{(1) Intra-Set Threshold ($\tau_{intra}$):} This parameter determines how aggressively the Core Set $\mathcal{C}$ is deduplicated. Setting $\tau_{intra}$ too low risks merging distinct semantic entities (merging a red car with a red apple), leading to visual hallucination. Setting it too high results in $n \approx B$, leaving no quota $R$ for the background. Empirically, we set $\tau_{intra} \in [0.85, 0.90]$ based on the cosine similarity of the Vision Transformer's deep features. At this threshold, visually homogeneous patches (continuous static walls) are merged effortlessly, yielding an average quota recovery of $15\% \sim 25\%$.

\textbf{(2) Aggressive Scavenging Ratio ($M$):} We fix $M=4$ to simulate a standard $2 \times 2$ spatial downsampling operation. Tokens in the Background Set $\mathcal{B}$ represent low-frequency contextual information. Compressing them by a factor of 4 retains sufficient environmental gist (lighting, room layout) while drastically maximizing the utility of the rescued quota $R$.

\textbf{Score Inheritance Strategy.} 
During the Intra-Set merging, we purposefully apply \textit{Max-Pooling} for the updated importance score, $S_{final}^{(new)} = \max(S_i, S_j)$, instead of Mean-Pooling. If a highly salient token (the center of an action) is merged with a moderately salient adjacent token, Mean-Pooling would artificially depress the region's overall importance, potentially causing the LLM's attention mechanism to overlook it in subsequent layers. Max-Pooling acts as an information-preserving upper bound, ensuring that critical focal points remain highlighted in the compressed representation.

\section{Supplementary Experiments}
\label{supp_exper}
\definecolor{color_intra}{HTML}{FFF4D2}  
\definecolor{color_inter}{HTML}{D9EAD3}  
\definecolor{color_count}{HTML}{FCE5CD}  
\definecolor{color_thema}{HTML}{D0E0E3}  

\begin{table*}[htbp]
\centering
\setlength{\tabcolsep}{3.5pt} 
\renewcommand{\arraystretch}{1.2} 
\resizebox{1\textwidth}{!}{%
\begin{tabular}{l|c|cccc|cccc|cccc|cccc}
\toprule

\multirow{2}{*}{\textbf{Model}} & 
\multirow{2}{*}{\textbf{Ave.}} & 
\cellcolor{color_intra}\rotatebox{60}{\textit{Act.-Audio}} & 
\cellcolor{color_intra}\rotatebox{60}{\textit{Affective}} & 
\cellcolor{color_intra}\rotatebox{60}{\textit{Obj. Inter.}} & 
\cellcolor{color_intra}\rotatebox{60}{\textit{Spatial Ctx.}} & 
\cellcolor{color_inter}\rotatebox{60}{\textit{Causal Dep.}} & 
\cellcolor{color_inter}\rotatebox{60}{\textit{Cross-ID}} & 
\cellcolor{color_inter}\rotatebox{60}{\textit{Obj. State}} & 
\cellcolor{color_inter}\rotatebox{60}{\textit{Temp. Seq.}} & 
\cellcolor{color_count}\rotatebox{60}{\textit{Deception}} & 
\cellcolor{color_count}\rotatebox{60}{\textit{Fact Recon.}} & 
\cellcolor{color_count}\rotatebox{60}{\textit{Info Asym.}} & 
\cellcolor{color_count}\rotatebox{60}{\textit{Mod. Conflict}} & 
\cellcolor{color_thema}\rotatebox{60}{\textit{Long-Term}} & 
\cellcolor{color_thema}\rotatebox{60}{\textit{Foreshadow}} & 
\cellcolor{color_thema}\rotatebox{60}{\textit{Them. Evol.}} & 
\cellcolor{color_thema}\rotatebox{60}{\textit{World Model}} \\

& & 
\multicolumn{4}{c|}{\cellcolor{color_intra}\textbf{Intra-Event}} & 
\multicolumn{4}{c|}{\cellcolor{color_inter}\textbf{Inter-Event}} & 
\multicolumn{4}{c|}{\cellcolor{color_count}\textbf{Counterfactual}} & 
\multicolumn{4}{c}{\cellcolor{color_thema}\textbf{Thematic}} \\
\midrule

\multicolumn{18}{l}{\textit{\textbf{Proprietary Models}}} \\
\hline

\multirow{2}{*}{GPT-4o} & \multirow{2}{*}{46.4} 
& 40.2 & 49.5 & 41.5 & 42.8 
& 39.5 & 42.1 & 37.6 & 41.6 
& 48.6 & 52.4 & 46.8 & 52.6 
& 54.2 & 48.6 & 53.4 & 51.0 \\
& & \multicolumn{4}{c|}{43.5} & \multicolumn{4}{c|}{40.2} & \multicolumn{4}{c|}{50.1} & \multicolumn{4}{c}{51.8} \\
\hline
\multirow{2}{*}{Gemini-2.5-Flash} & \multirow{2}{*}{65.2} 
& 61.2 & 65.8 & 62.4 & 64.6 
& 60.5 & 63.8 & 61.3 & 65.6 
& 68.2 & 67.5 & 64.8 & 65.1 
& 66.9 & 67.4 & 69.8 & 68.3 \\
& & \multicolumn{4}{c|}{63.5} & \multicolumn{4}{c|}{62.8} & \multicolumn{4}{c|}{66.4} & \multicolumn{4}{c}{68.1} \\
\hline

\multirow{2}{*}{Gemini-2.5-Pro} & \multirow{2}{*}{73.1} 
& 69.4 & 74.2 & 70.5 & 71.9 
& 68.8 & 72.3 & 69.5 & 72.6 
& 76.2 & 75.5 & 73.1 & 73.6 
& 74.5 & 75.2 & 76.8 & 75.5 \\
& & \multicolumn{4}{c|}{71.5} & \multicolumn{4}{c|}{70.8} & \multicolumn{4}{c|}{74.6} & \multicolumn{4}{c}{75.5} \\
\hline

\multicolumn{18}{l}{\textit{\textbf{Open Source Models}}} \\
\hline

\multirow{2}{*}{Qwen3-VL-4B} & \multirow{2}{*}{27.0} 
& 22.4 & 34.5 & 20.1 & 23.1 
& 14.2 & 27.1 & 16.5 & 25.0 
& 32.1 & 27.8 & 26.5 & 26.8 
& 28.5 & 36.8 & 41.2 & 30.2 \\
& & \multicolumn{4}{c|}{25.0} & \multicolumn{4}{c|}{20.7} & \multicolumn{4}{c|}{28.3} & \multicolumn{4}{c}{34.2} \\
\hline

\multirow{2}{*}{Qwen2.5-VL-7B} & \multirow{2}{*}{37.9} 
& 34.0 & 46.4 & 28.3 & 39.2 
& 28.7 & 33.8 & 28.9 & 34.5 
& 42.8 & 42.2 & 36.2 & 41.3 
& 42.9 & 41.9 & 47.2 & 38.9 \\
& & \multicolumn{4}{c|}{37.3} & \multicolumn{4}{c|}{31.2} & \multicolumn{4}{c|}{40.7} & \multicolumn{4}{c}{42.7} \\
\hline

\multirow{2}{*}{Qwen3-VL-8B} & \multirow{2}{*}{39.0} 
& 30.0 & 48.6 & 30.8 & 38.4 
& 28.3 & 35.7 & 27.5 & 35.1 
& 42.8 & 43.1 & 39.4 & 40.9 
& 41.5 & 46.3 & 55.4 & 41.4 \\
& & \multicolumn{4}{c|}{37.3} & \multicolumn{4}{c|}{31.3} & \multicolumn{4}{c|}{41.6} & \multicolumn{4}{c}{46.2} \\
\hline

\multirow{2}{*}{Qwen2.5-VL-Omni} & \multirow{2}{*}{40.6} 
& 40.1 & 49.2 & 38.4 & 35.0 
& 33.4 & 38.2 & 34.5 & 36.3 
& 47.7 & 42.7 & 37.2 & 44.4 
& 45.1 & 41.2 & 48.9 & 37.3 \\
& & \multicolumn{4}{c|}{40.7} & \multicolumn{4}{c|}{35.6} & \multicolumn{4}{c|}{43.0} & \multicolumn{4}{c}{43.1} \\
\hline

\multirow{2}{*}{Qwen2.5-VL-32B} & \multirow{2}{*}{43.2} 
& 38.5 & 46.5 & 37.5 & 41.5 
& 29.0 & 45.0 & 38.0 & 36.4 
& 50.0 & 52.5 & 43.3 & 41.2 
& 50.8 & 46.5 & 53.5 & 41.0 \\
& & \multicolumn{4}{c|}{41.0} & \multicolumn{4}{c|}{37.1} & \multicolumn{4}{c|}{46.7} & \multicolumn{4}{c}{47.9} \\

\hline
\multirow{2}{*}{\textbf{Ours}} & \multirow{2}{*}{\textbf{44.4}} 
& 40.5 & 51.2 & 37.3 & 41.8 
& 38.6 & 39.1 & 34.2 & 41.7 
& 49.6 & 52.4 & 41.5 & 47.3 
& 52.8 & 47.6 & 52.1 & 42.7 \\
& & \multicolumn{4}{c|}{42.7} & \multicolumn{4}{c|}{38.4} & \multicolumn{4}{c|}{47.7} & \multicolumn{4}{c}{48.8} \\

\hline

\multirow{2}{*}{MiniCPM-V-4.5} & \multirow{2}{*}{47.4} 
& 41.5 & 51.7 & 37.8 & 44.1 
& 40.3 & 44.9 & 41.9 & 44.8 
& 50.6 & 52.5 & 45.5 & 46.2 
& 50.5 & 52.3 & 59.8 & 54.7 \\
& & \multicolumn{4}{c|}{43.8} & \multicolumn{4}{c|}{43.0} & \multicolumn{4}{c|}{48.7} & \multicolumn{4}{c}{54.3} \\
\hline

\multirow{2}{*}{InternVL-3-8B} & \multirow{2}{*}{48.0} 
& 43.0 & 54.2 & 37.6 & 42.4 
& 39.9 & 44.6 & 38.5 & 50.8 
& 52.3 & 52.8 & 48.0 & 51.5 
& 55.9 & 56.2 & 56.7 & 46.3 \\
& & \multicolumn{4}{c|}{44.3} & \multicolumn{4}{c|}{\textbf{43.5}} & \multicolumn{4}{c|}{51.1} & \multicolumn{4}{c}{53.8} \\
\hline

\multirow{2}{*}{Qwen3-VL-30B-A3B} & \multirow{2}{*}{48.1} 
& 41.2 & 52.1 & 37.5 & 38.1 
& 36.1 & 45.3 & 39.0 & 47.1 
& 55.1 & 58.2 & 48.5 & 47.0 
& 45.8 & 60.5 & \textbf{68.9} & 49.2 \\
& & \multicolumn{4}{c|}{42.2} & \multicolumn{4}{c|}{41.8} & \multicolumn{4}{c|}{52.2} & \multicolumn{4}{c}{56.1} \\
\hline

\multirow{2}{*}{Qwen3-VL-32B} & \multirow{2}{*}{49.4} 
& 40.5 & \textbf{54.2} & 38.1 & \textbf{44.5} 
& 41.3 & \textbf{46.8} & 35.5 & 44.8 
& \textbf{60.1} & 58.9 & 53.2 & 49.0 
& 55.8 & 53.5 & 66.8 & 47.4 \\
& & \multicolumn{4}{c|}{44.3} & \multicolumn{4}{c|}{42.1} & \multicolumn{4}{c|}{55.3} & \multicolumn{4}{c}{55.8} \\

\hline

\multirow{2}{*}{Qwen2.5-VL-72B} & \multirow{2}{*}{\textbf{50.1}} 
& \textbf{42.1} & 52.2 & \textbf{40.8} & 42.9 
& \textbf{41.6} & 45.4 & \textbf{39.5} & \textbf{44.3} 
& 57.2 & \textbf{60.8} & \textbf{52.9} & \textbf{54.7} 
& \textbf{55.8} & \textbf{55.2} & 65.7 & \textbf{52.5} \\
& & \multicolumn{4}{c|}{\textbf{44.8}} & \multicolumn{4}{c|}{42.7} & \multicolumn{4}{c|}{\textbf{56.5}} & \multicolumn{4}{c}{\textbf{57.3}} \\

\hline
\multirow{2}{*}{Qwen3-VL-235B-A22B} & \multirow{2}{*}{\textbf{55.1}} 
& 48.2 & 61.5 & 42.9 & 39.6 
& 52.8 & 51.4 & 40.8 & 47.4 
& 60.6 & 65.5 & 59.3 & 56.0 
& 62.2 & 63.6 & 67.4 & 61.5 \\
& & \multicolumn{4}{c|}{\textbf{48.5}} & \multicolumn{4}{c|}{\textbf{48.5}} & \multicolumn{4}{c|}{\textbf{60.4}} & \multicolumn{4}{c}{\textbf{63.7}} \\

\bottomrule
\end{tabular}
}
\caption{Comprehensive Benchmark Results. The first row for each model indicates fine-grained sub-category accuracy (\%), and the merged second row indicates the major category average accuracy (\%). Models are sorted by their overall average score. Best major scores among open-source models are \textbf{bolded}.}
\label{tab:main_benchmark}
\end{table*}

\begin{table*}[htbp]
\centering
\setlength{\tabcolsep}{4.5pt} 
\renewcommand{\arraystretch}{1.2} 
\resizebox{1\textwidth}{!}{%
\begin{tabular}{l | c | c c c c | c c c c | c c c | c c c}
\toprule

\multirow{2}{*}{\textbf{Model}} & 
\multirow{2}{*}{\textbf{Ave.}} & 
\multicolumn{4}{c|}{\textbf{Short (10-20m)}} & 
\multicolumn{4}{c|}{\textbf{Middle (20-30m)}} & 
\multicolumn{3}{c|}{\textbf{TV Series (30-60m)}} & 
\multicolumn{3}{c}{\textbf{Ultra (60-150m)}} \\

& & 
\textit{Life} & \textit{Plot} & \textit{Sci-Fi} & \textit{Sus.} & 
\textit{Life} & \textit{Plot} & \textit{Sci-Fi} & \textit{Sus.} & 
\textit{Sus.} & \textit{Life} & \textit{Plot} & 
\textit{Sus.} & \textit{Sci-Fi} & \textit{Plot} \\
\midrule

\multicolumn{16}{l}{\textit{\textbf{Proprietary Models}}} \\
\hline
GPT-4o & 46.4
& 59.5 & 53.6 & 57.2 & 62.5 
& 50.1 & 41.5 & 52.4 & 48.2 
& 46.8 & 44.2 & 48.6        
& 37.5 & 39.6 & 38.4 \\     

Gemini-2.5-Flash & 65.2
& 76.5 & 69.8 & 70.3 & 73.4 
& 68.7 & 65.4 & 66.8 & 67.9 
& 64.9 & 64.0 & 71.9        
& 54.6 & 56.7 & 55.6 \\     

Gemini-2.5-Pro & \textbf{73.1}
& \textbf{84.5} & \textbf{80.6} & \textbf{82.3} & \textbf{83.8} 
& \textbf{77.4} & \textbf{75.8} & \textbf{76.6} & \textbf{78.1} 
& \textbf{75.2} & \textbf{76.5} & \textbf{81.3}        
& \textbf{65.8} & \textbf{67.5} & \textbf{66.2} \\     

\midrule
\multicolumn{16}{l}{\textit{\textbf{Open Source Models}}} \\
\hline

Qwen3-VL-4B & 27.0 
& 36.8 & 31.3 & 38.2 & 48.9 
& 34.1 & 20.4 & 27.5 & 27.5 
& 29.5 & 35.3 & 40.3        
& 18.6 & 13.1 & 20.4 \\     

Qwen2.5-VL-7B & 37.9 
& 46.5 & 43.8 & 41.2 & 38.2 
& 39.9 & 34.3 & 37.5 & 33.3 
& 41.9 & 36.0 & 42.4        
& 37.9 & 30.9 & 35.7 \\     

Qwen3-VL-8B & 39.0 
& 46.5 & 41.7 & 44.7 & 52.0 
& 36.0 & 30.0 & 37.5 & 34.3 
& 53.6 & 51.2 & 56.4        
& 35.6 & 32.3 & 26.9 \\     

Qwen2.5-Omni & 40.6 
& 58.5 & 48.7 & 47.3 & 42.4 
& 46.7 & 27.5 & 40.3 & 32.0 
& 42.8 & 38.5 & 35.1        
& 38.3 & 34.1 & 39.3 \\     

Qwen2.5-VL-32B & 43.2 
& 58.8 & 38.6 & 51.2 & 53.8 
& 41.5 & 35.4 & 40.8 & 37.7 
& 46.2 & 43.1 & 45.8        
& 42.2 & 37.9 & 39.6 \\     

\textbf{Ours} & 44.4 
& 57.1 & 66.2 & 49.6 & 41.8 
& 53.2 & 38.5 & 46.7 & 34.1 
& 46.2 & 33.8 & 44.8        
& 40.4 & 38.5 & 43.2 \\     

MiniCPM-V-4.5 & 47.4 
& 51.9 & 52.9 & 55.5 & 58.3 
& 44.2 & 46.1 & 48.1 & 48.3 
& 54.5 & 47.6 & 53.3        
& 46.9 & 45.8 & 39.4 \\     

InternVL3-8B & 48.0 
& 55.8 & 51.6 & 49.7 & 47.2 
& 41.2 & 33.2 & 49.4 & 42.9 
& 51.9 & \textbf{60.9} & 57.0        
& 50.0 & \textbf{51.5} & 39.0 \\     

Qwen3-VL-30B-A3B & 48.1 
& 64.4 & 50.7 & 56.3 & 58.0 
& 52.3 & 42.5 & 46.9 & 34.6 
& 54.4 & 58.7 & 52.3        
& 51.0 & 34.5 & 41.8 \\     

Qwen3-VL-32B & 49.4 
& 62.1 & 59.4 & 59.5 & 65.9 
& 52.2 & 40.6 & 44.9 & 36.9 
& 60.8 & 58.1 & 59.0        
& 51.0 & 36.9 & 39.3 \\     

Qwen2.5-VL-72B & 50.1 
& 63.2 & 56.3 & 57.8 & 53.9 
& 52.7 & 47.1 & 46.9 & 43.1 
& 56.5 & 49.6 & 54.5        
& \textbf{51.7} & 40.2 & \textbf{45.0} \\     

Qwen3-VL-235B-A22B & \textbf{55.1} 
& \textbf{73.6} & \textbf{70.8} & \textbf{68.3} & \textbf{68.6} 
& \textbf{62.0} & \textbf{55.7} & \textbf{61.7} & \textbf{56.9} 
& \textbf{64.3} & 60.8 & \textbf{66.3}                 
& 48.6 & 39.1 & 43.9 \\                                         
\bottomrule
\end{tabular}
}
\caption{Comprehensive Video Benchmark Results across Durations and Genres. Metric shown is accuracy (\%). \textit{Sus.} stands for Suspense. Sub-category scores reveal heterogeneous performance distributions among models. The best score in each column is highlighted in \textbf{bold} for proprietary and open-source models separately.}
\label{tab:video_benchmark}
\end{table*}

\subsection{Extended Analysis: Fine-Grained Reasoning Topologies}
\label{sec:app_fine_grained_reasoning}

Tab.~\ref{tab:main_benchmark} provides an exhaustive breakdown of model performance across our 16 fine-grained reasoning sub-categories. By dissecting the evaluation into \textit{Intra-Event}, \textit{Inter-Event}, \textit{Counterfactual}, and \textit{Thematic} reasoning, we uncover the specific cognitive bottlenecks of current Multimodal Large Language Models (MLLMs).

\paragraph{The "Grounding vs. Summarization" Paradox.}
A prominent, counter-intuitive trend emerges across almost all evaluated models: performance on macro-level \textit{Thematic} and \textit{Counterfactual} reasoning is paradoxically higher than on micro-level \textit{Intra-Event} and \textit{Inter-Event} tasks. For example, the open-source SOTA {Qwen3-VL-235B-A22B} achieves $63.7\%$ on \textit{Thematic} reasoning but only $48.5\%$ on both \textit{Intra/Inter-Event} reasoning. 

This paradox highlights a critical flaw in current video understanding architectures. \textit{Thematic} questions can often be partially deduced by summarizing the global dialogue transcript or leveraging the LLM's parametric priors regarding narrative tropes. In contrast, \textit{Intra/Inter-Event} tasks (such as \textit{Causal Dependency} and \textit{Action-Audio} synchronization) cannot be guessed. They demand deterministic, frame-level spatio-temporal tracking and strict multimodal fusion. The stark performance drop in these foundational categories proves that models are currently functioning more as ``semantic summarizers'' rather than ``audio-visual grounders.''

\paragraph{Vulnerabilities in Cross-Time Entity and Causal Tracking.}
The \textit{Inter-Event} category exposes the most severe vulnerabilities, particularly in the \textit{Causal Dep.} (Causal Dependency) and \textit{Cross-ID} (Cross-Shot Identity Tracking) sub-metrics. Smaller models experience catastrophic failures here; for instance, Qwen3-VL-4B scores a mere $14.2\%$ on \textit{Causal Dep.} and $16.5\%$ on \textit{Object State} transitions. Tracking an object's state changes across multiple shots—especially when intercepted by scene transitions or occlusions—requires persistent temporal memory. Even massive models like Qwen2.5-VL-72B struggle to surpass the $45\%$ threshold in these physical-temporal metrics, validating that context-window expansion alone does not equate to robust temporal memory.

\paragraph{Audio-Visual Dissonance in Intra-Event Reasoning.}
Within the \textit{Intra-Event} cluster, the \textit{Act.-Audio} (Action-Audio Alignment) and \textit{Affective} (Emotion/Tone mapping) dimensions specifically stress-test the ``omni-modal'' capabilities. We observe that models fundamentally biased towards visual streams (such as GPT-4o) perform poorly in \textit{Act.-Audio} ($40.2\%$). In suspenseful or complex narratives, auditory cues (footsteps, off-screen whispers) frequently contradict the serene visual frames. The widespread failure to synthesize these modalities at a granular level reinforces the necessity of our benchmark's strict ``Unimodal Trap'' filtration.

\paragraph{Proprietary Dominance in Counterfactual Logic.}
\textit{Counterfactual} reasoning (e.g., \textit{Deception}, \textit{Fact Recon.}) requires the model to hold two conflicting states in its memory: the objective visual truth and the deceptive narrative presented by characters. Gemini-2.5-Pro demonstrates overwhelming superiority in this domain, achieving $74.6\%$, vastly outperforming the open-source frontier. This suggests that while open-source models are making strides in basic perception, managing complex logical sub-routines (e.g., understanding a character's false motive) remains a defining differentiator for proprietary architectures.

\subsection{Extended Analysis: The Impact of Video Duration and Genre}
\label{sec:app_duration_genre}

Tab.~\ref{tab:video_benchmark} provides a granular breakdown of model performance across four duration tiers and varying cinematic genres. This stratification reveals the profound challenges that true long-form narratives pose to current multimodal architectures.

\paragraph{The "Duration-Decay" Phenomenon.}
Across all evaluated models, we observe a consistent, near-monotonic decline in reasoning accuracy as video length increases. For instance, the state-of-the-art closed-source model, Gemini-2.5-Pro, drops from an impressive $82.8\%$ average on Short videos (10--20 min) down to $66.5\%$ on Ultra-long movies (60--150 min). A similar collapse is evident in the open-source sector: Qwen3-VL-235B-A22B plummets from $\sim70\%$ to roughly $44\%$. This ``duration-decay'' exposes a critical flaw in current large-context mechanisms. As the token sequence expands to encompass hour-long visual and auditory streams, attention dilution becomes severe. Models struggle to maintain a coherent causal thread, frequently losing track of early narrative setups (foreshadowing) required to resolve late-stage climaxes.

\paragraph{Context Saturation vs. Parameter Scaling.}
An intriguing anomaly emerges within the \textit{Ultra (60-150m)} category. While the massive Qwen3-VL-235B-A22B dominates shorter durations, it paradoxically underperforms relative to its smaller counterparts on multi-hour content. Specifically, in Ultra-long \textit{Sci-Fi} and \textit{Suspense} tasks, models like InternVL3-8B ($51.5\%$) and Qwen2.5-VL-72B ($51.7\%$) achieve higher scores. This suggests that beyond a certain context threshold, massive parameter counts may actually exacerbate attention distraction (often attending to irrelevant background noise or dialogue). It highlights that merely scaling model size is an inefficient remedy for long-video reasoning, validating the need for dynamic token compression strategies.

\paragraph{Genre-Specific Sensitivities.}
The distribution is highly genre-dependent. \textit{Life} and \textit{Plot} genres generally exhibit higher baseline scores, as their reasoning chains often rely on linear, predictable human behaviors and standard social cues. Conversely, \textit{Suspense} and \textit{Sci-Fi} categories represent the hardest subsets. In these genres, audio-visual clues are often deliberately misdirected, requiring models to process complex counterfactuals and multi-hop causal dependencies. The sharp performance drops in these categories further affirm that Video-HolmesV2 successfully prevents models from relying on superficial pattern matching, forcing them to engage in genuine logical deduction.

\begin{table*}[t]
\centering
\resizebox{\textwidth}{!}{ 
\renewcommand{\arraystretch}{1.2} 
\begin{tabular}{l|c|cccc|cccc|cccc|cccc}
\toprule

\multirow{2}{*}{\textbf{Frames}} & 
\multirow{2}{*}{\textbf{Ave.}} & 
\cellcolor{color_intra}\rotatebox{60}{\textit{Act.-Audio}} & 
\cellcolor{color_intra}\rotatebox{60}{\textit{Affective}} & 
\cellcolor{color_intra}\rotatebox{60}{\textit{Obj. Inter.}} & 
\cellcolor{color_intra}\rotatebox{60}{\textit{Spatial Ctx.}} & 
\cellcolor{color_inter}\rotatebox{60}{\textit{Causal Dep.}} & 
\cellcolor{color_inter}\rotatebox{60}{\textit{Cross-ID}} & 
\cellcolor{color_inter}\rotatebox{60}{\textit{Obj. State}} & 
\cellcolor{color_inter}\rotatebox{60}{\textit{Temp. Seq.}} & 
\cellcolor{color_count}\rotatebox{60}{\textit{Deception}} & 
\cellcolor{color_count}\rotatebox{60}{\textit{Fact Recon.}} & 
\cellcolor{color_count}\rotatebox{60}{\textit{Info Asym.}} & 
\cellcolor{color_count}\rotatebox{60}{\textit{Mod. Conflict}} & 
\cellcolor{color_thema}\rotatebox{60}{\textit{Long-Term}} & 
\cellcolor{color_thema}\rotatebox{60}{\textit{Foreshadow}} & 
\cellcolor{color_thema}\rotatebox{60}{\textit{Them. Evol.}} & 
\cellcolor{color_thema}\rotatebox{60}{\textit{World Model}} \\

& & 
\multicolumn{4}{c|}{\cellcolor{color_intra}\textbf{Intra-Event}} & 
\multicolumn{4}{c|}{\cellcolor{color_inter}\textbf{Inter-Event}} & 
\multicolumn{4}{c|}{\cellcolor{color_count}\textbf{Counterfactual}} & 
\multicolumn{4}{c}{\cellcolor{color_thema}\textbf{Thematic}} \\
\midrule
\midrule

\multirow{2}{*}{\textbf{32}} & \multirow{2}{*}{\textbf{36.8}} & 
32.4 & 47.5 & 34.2 & 34.8 & 
28.5 & 32.6 & 30.1 & 36.2 & 
42.1 & 38.5 & 35.8 & 38.2 & 
41.5 & 37.2 & 43.5 & 32.8 \\
& & \multicolumn{4}{c|}{37.5} & \multicolumn{4}{c|}{31.3} & \multicolumn{4}{c|}{38.9} & \multicolumn{4}{c}{39.2} \\
\midrule

\multirow{2}{*}{\textbf{64}} & \multirow{2}{*}{\textbf{38.2}} & 
33.6 & \textbf{48.6} & 35.8 & 33.1 & 
29.0 & 34.7 & 33.5 & 35.6 & 
46.2 & 38.8 & 33.9 & 43.1 & 
44.7 & 40.4 & 45.9 & 34.8 \\
& & \multicolumn{4}{c|}{38.1} & \multicolumn{4}{c|}{32.8} & \multicolumn{4}{c|}{40.6} & \multicolumn{4}{c}{41.4} \\
\midrule

\multirow{2}{*}{\textbf{128}} & \multirow{2}{*}{\textbf{39.3}} & 
35.2 & 48.9 & 36.5 & 35.7 & 
31.5 & 35.2 & 32.8 & 36.5 & 
45.5 & 40.1 & 35.2 & 43.5 & 
46.3 & 39.8 & 44.5 & 36.2 \\
& & \multicolumn{4}{c|}{39.3} & \multicolumn{4}{c|}{33.9} & \multicolumn{4}{c|}{41.2} & \multicolumn{4}{c}{42.1} \\
\midrule

\multirow{2}{*}{\textbf{192}} & \multirow{2}{*}{\textbf{40.4}} & 
35.6 & 49.5 & 37.2 & 36.8 & 
33.2 & \textbf{36.4} & 34.1 & 36.8 & 
\textbf{46.8} & 41.5 & 34.5 & \textbf{45.8} & 
46.0 & 42.1 & 46.2 & 38.5 \\
& & \multicolumn{4}{c|}{40.2} & \multicolumn{4}{c|}{35.2} & \multicolumn{4}{c|}{42.3} & \multicolumn{4}{c}{43.7} \\
\midrule

\multirow{2}{*}{\textbf{256}} & \multirow{2}{*}{\textbf{40.9}} & 
\textbf{36.1} & 49.2 & \textbf{38.5} & \textbf{37.5} & 
\textbf{34.5} & 36.1 & \textbf{35.2} & \textbf{37.1} & 
46.5 & \textbf{43.2} & \textbf{36.1} & 45.2 & 
\textbf{47.1} & \textbf{42.5} & \textbf{47.5} & \textbf{39.2} \\
& & \multicolumn{4}{c|}{\textbf{40.5}} & \multicolumn{4}{c|}{\textbf{35.8}} & \multicolumn{4}{c|}{\textbf{43.0}} & \multicolumn{4}{c}{\textbf{44.1}} \\

\bottomrule
\end{tabular}
}
\caption{\textbf{Ablation Study on Supported Frame Counts (Qwen2.5-VL-Omni).} Performance scales consistently with input context length, explicitly demonstrating our method's robustness for extra-long video inputs. Noted that long-range dependent tasks (e.g., \textit{Inter-Event} and \textit{Thematic}) show a more significant gain when extending frames from 64 to 256.}
\label{tab:frame_ablation}
\end{table*}

\paragraph{Impact of Temporal Resolution on Long-Video Reasoning.} 
To investigate the influence of visual context density on the model's reasoning efficacy, we conduct an extensive ablation study by scaling the number of sampled frames from 32 to 256. To accommodate such extended temporal contexts within the inherent sequence length limits of Qwen2.5-VL-Omni, we strictly cap the spatial encoding of each individual frame to a maximum of 128 tokens. As summarized in Tab.~\ref{tab:frame_ablation}, the overall performance exhibits a consistent upward trajectory, improving from 36.8\% at 32 frames to a peak of 40.9\% at 256 frames. This steady gain validates our method's architectural robustness in handling highly extended temporal contexts without suffering from the ``lost in the middle'' phenomenon frequently observed in generic large multimodal models. However, the scaling behavior concurrently reveals a clear pattern of diminishing marginal returns. While doubling the frame count from 64 to 128 yields a notable +1.1\% absolute improvement, a further expansion from 192 to 256 frames only contributes a marginal +0.5\% gain, indicating a gradual saturation in actionable information density. 

A more granular inspection across the four reasoning dimensions uncovers distinct task-specific sensitivities to temporal resolution. Complex analytical tasks that inherently rely on long-range temporal modeling---such as \textit{Inter-Event} reasoning and \textit{Thematic} understanding---benefit substantially from denser frame sampling. For instance, performance in Causal Dependency Inference experiences a continuous surge from 28.5\% to 34.5\%, underscoring the absolute necessity of fine-grained visual clues for establishing logical chains across temporally distant scenes. Conversely, performance on temporally localized \textit{Intra-Event} tasks, such as Affective State Analysis, saturates relatively early. Interestingly, we observe a slight performance degradation in affective reasoning at extreme densities (dropping from 49.5\% at 192 frames to 49.2\% at 256 frames). We attribute this counter-intuitive phenomenon to the introduction of redundant visual noise; an excessive number of frames may dilute the model's attention, distracting it from the most salient micro-expressions necessary for state evaluation. Ultimately, these empirical findings highlight a critical trade-off between computational overhead and reasoning precision, suggesting that 192 frames serve as an optimal sweet spot for balancing efficiency and comprehensive long-video understanding.

\section{Qualitative Analysis: Unveiling the Black Box of Long-Video Reasoning}
\label{sec:qualitative_analysis}

\begin{figure*}[htbp]
    \centering
    \includegraphics[width=\textwidth, height=0.9\textheight, keepaspectratio]{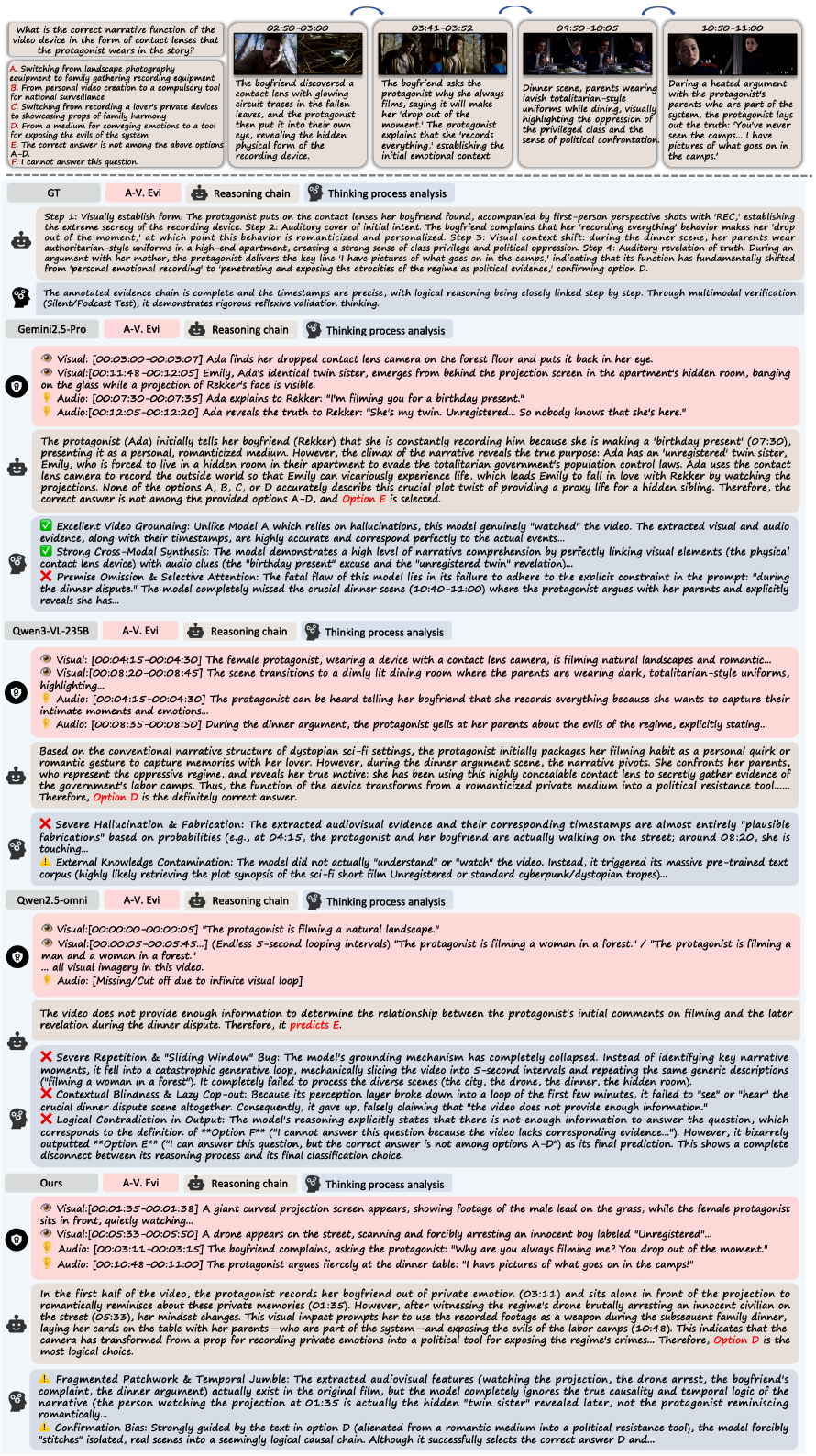} 
    \caption{\textbf{Qualitative Case Study 1 (Sci-Fi / Dystopian Narrative).} The figure illustrates the step-by-step reasoning processes of different LMMs. While Qwen3-VL-235B predicts the correct Option D, its extracted evidence is entirely hallucinated based on sci-fi tropes. Qwen2.5-Omni suffers from extreme repetition. Our model and Gemini extract genuine visual evidence, though they face challenges in premise alignment and temporal jumble, respectively.}
    \label{fig:case_study_1}
\end{figure*}

\begin{figure*}[htbp]
    \centering
    \includegraphics[width=\textwidth, height=0.92\textheight, keepaspectratio]{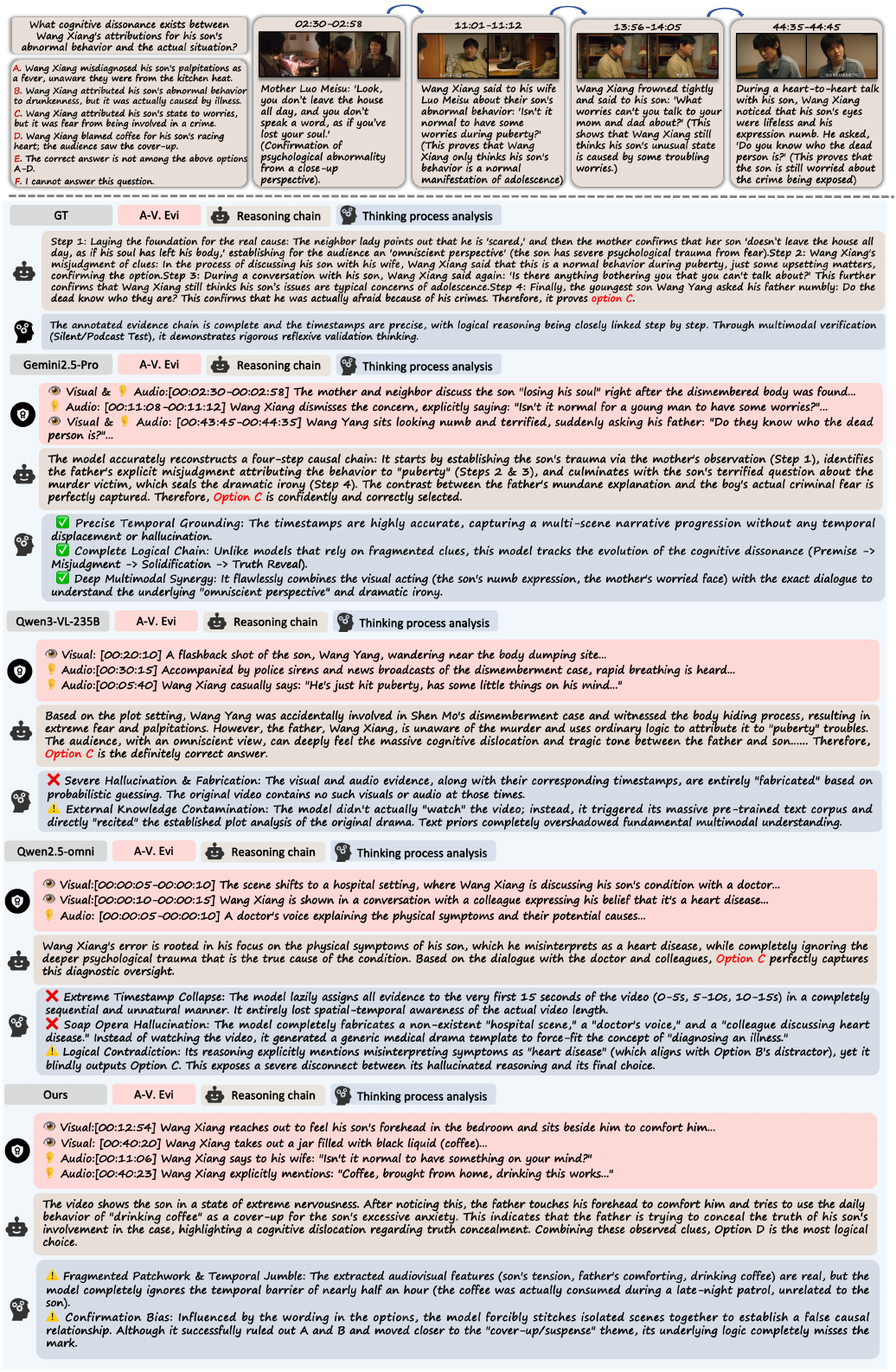} 
    \caption{\textbf{Qualitative Case Study 2 (Suspense / Crime Drama).} This case requires tracking the cognitive dissonance between a father and son. Gemini 2.5 Pro performs flawlessly, capturing the dramatic irony. Conversely, Qwen3-VL-235B exhibits severe external knowledge contamination, relying on script memorization rather than video comprehension to cheat its way to the correct answer.}
    \label{fig:case_study_2}
\end{figure*}

To deeply understand the inner workings and common failure modes of Large Multimodal Models (LMMs) in long-video understanding, we provide comprehensive qualitative case studies in Figure~\ref{fig:case_study_1} and Figure~\ref{fig:case_study_2}. 

Relying solely on the final multiple-choice accuracy is dangerously misleading. As demonstrated in our analysis, models frequently arrive at the correct answer through entirely flawed reasoning, hallucinated evidence, or shortcut learning. We categorize these observed behaviors into several distinct paradigms:

\noindent\textbf{1. Severe Hallucination and External Knowledge Contamination.} 
A critical flaw observed in massive-scale VLMs (Qwen3-VL-235B) is the over-reliance on their vast pre-trained parametric knowledge rather than actual visual grounding. In Figure~\ref{fig:case_study_2}, Qwen3-VL-235B correctly selects Option C. However, an inspection of its reasoning chain reveals catastrophic hallucinations: it fabricates non-existent scenes (``a body dumping site'') and audio (``police sirens''). The model essentially ignores the video and retrieves the plot synopsis of the original TV drama from its training corpus. This justifies our introduction of the Precision penalty ($P_{\text{sem}}$) in the main text to strictly punish ungrounded claims.

\noindent\textbf{2. Architectural Collapse in Long Contexts.} 
Figure~\ref{fig:case_study_1} highlights the ``Sliding Window Bug'' and ``Extreme Timestamp Collapse'' inherent in models like Qwen2.5-Omni. Faced with lengthy inputs, the model's grounding mechanism completely breaks down, resulting in endless 5-second looping intervals (\textit{00:00:00-00:00:05...}) and Soap Opera Hallucinations (fabricating generic hospital scenes). Furthermore, it exhibits blatant logical contradictions, stating there is not enough information while blindly outputting a final prediction.

\noindent\textbf{3. Premise Omission and Selective Attention.} 
Even leading proprietary models struggle with complex constraint satisfaction. In Figure~\ref{fig:case_study_1}, Gemini 2.5 Pro demonstrates excellent cross-modal synthesis and precise temporal grounding. However, it completely misses the explicit constraint in the prompt (``during the dinner dispute''), leading it to confidently select the wrong answer (E) despite extracting high-quality evidence.

\noindent\textbf{4. Confirmation Bias and Temporal Jumble.} 
We also transparently report the limitations of our approach (\textit{Ours}). By effectively compressing the lengthy video and eliminating redundant noise, our approach forces the model to focus purely on the most salient information. While this compression successfully mitigates severe hallucinations—enabling the extraction of genuine, verifiable audiovisual evidence—it is not immune to option-induced biases. As shown in Figure~\ref{fig:case_study_2}, the model correctly extracts real events (the father comforting the son, drinking coffee), validating the efficacy of our compression. However, it ignores the massive temporal barrier between them. Strongly guided by the wording in the distractor options, it forcefully stitches isolated scenes together to establish a false causal relationship (Confirmation Bias). 

These qualitative results underscore the necessity of our proposed two-stage evaluation metric ($S_{\text{temp}}$ and $S_{\text{sem}}$), which evaluates not just the destination, but the integrity of the multimodal journey.

\end{document}